\documentclass[journal]{IEEEtran}
\makeatletter

\newcommand{\Rnum}[1]{\expandafter\@slowromancap\romannumeral #1@}
\usepackage{eufrak}
\usepackage{hyperref}
\usepackage{ifpdf}

\usepackage{multirow}

\usepackage{cite}

\ifCLASSINFOpdf
   \usepackage[pdftex]{graphicx}
   \graphicspath{{../pdf/}{../jpeg/}}
   \DeclareGraphicsExtensions{.pdf,.jpeg,.png}
\else
   \usepackage[dvips]{graphicx}
   \graphicspath{{../eps/}}
   \DeclareGraphicsExtensions{.eps}
\fi
\usepackage{amsmath}
\usepackage{algorithmic}

\usepackage{array}

\ifCLASSOPTIONcompsoc
  \usepackage[caption=false,font=normalsize,labelfont=sf,textfont=sf]{subfig}
\else
  \usepackage[caption=false,font=footnotesize]{subfig}
\fi
\usepackage{url}

\usepackage{color}

\begin{document}

\title{
Low-Rank Pairwise Alignment Bilinear Network For Few-Shot Fine-Grained Image Classification}

\author{Huaxi Huang,~\IEEEmembership{Student Member,~IEEE,}
        Junjie Zhang,~
        Jian Zhang,~\IEEEmembership{Senior Member,~IEEE,}\\
        Jingsong Xu,~
        Qiang Wu,~\IEEEmembership{Senior Member,~IEEE}

\thanks{
Huaxi Huang and Junjie Zhang are co-first authors. Corresponding author: Jian Zhang, email: Jian.Zhang@uts.edu.au.

Huaxi Huang, Jian Zhang, Qiang Wu and Jingsong Xu are with the Faculty of Engineering and Information Technology, University of Technology Sydney, Sydney NSW 2007, Australia.
Junjie Zhang is with the School of Computer Science, The University of Adelaide, Adelaide SA 5005, Australia.

The preliminary version of this work is accepted at IEEE ICME 2019\cite{huang2019compare}.}
}

\markboth{Journal of \LaTeX\ Class Files,~Vol.~XX, No.~X, June~2019}%
{Shell \MakeLowercase{\textit{et al.}}: Bare Demo of IEEEtran.cls for IEEE Journals}

\maketitle

\begin{abstract}
Deep neural networks have demonstrated advanced abilities on various visual classification tasks, which heavily rely on the large-scale training samples with annotated ground-truth. However, it is unrealistic always to require such annotation in real-world applications. Recently, Few-Shot learning (FS), as an attempt to address the shortage of training samples, has made significant progress in generic classification tasks. Nonetheless, it is still challenging for current FS models to distinguish the subtle differences between fine-grained categories given limited training data. 
To filling the classification gap, in this paper, we address the Few-Shot Fine-Grained (FSFG) classification problem, which focuses on tackling the fine-grained classification under the challenging few-shot learning setting. A novel low-rank pairwise bilinear pooling operation is proposed to capture the nuanced differences between the support and query images for learning an effective distance metric. Moreover, a feature alignment layer is designed to match the support image features with query ones before the comparison. We name the proposed model Low-Rank Pairwise Alignment Bilinear Network (LRPABN), which is trained in an end-to-end fashion. Comprehensive experimental results on four widely used fine-grained classification data sets demonstrate that our LRPABN model achieves the superior performances compared to state-of-the-art methods.

\end{abstract}

\begin{IEEEkeywords}
	Few-Shot, Fine-Grained, Low-Rank, Pairwise, Bilinear Pooling, Feature Alignment.
\end{IEEEkeywords}

\IEEEpeerreviewmaketitle

\section{Introduction}
\label{sec:intro}

\IEEEPARstart{F}{ine-grained} 
image classification aims to distinguish different sub-categories belong to the same entry-level category such as birds~\cite{WahCUB_200_2011,Horn_2015_CVPR}, dogs~\cite{KhoslaYaoJayadevaprakashFeiFei_FGVC2011}, and cars~\cite{KrauseStarkDengFei-Fei_3DRR2013}. This problem is particularly challenging due to the low inter-category variance yet high intra-category discordance caused by various object postures, illumination conditions and distances from the cameras, \textit{etc.}  
In general, the majority of fine-grained classification approaches need to be fed with a large amount of training data before obtaining a trustworthy classifier~\cite{zhang2014part,Fu_2017_CVPR,Lin_2015_ICCV,Cui_2017_CVPR,Li_2018_CVPR,Krause_2015_CVPR}. However, labeling the fine-grained data requires strong domain knowledge, \textit{e.g.}, only ornithologists can accurately identify different bird species, which is significantly expensive compared to the generic object classification task. Moreover, in some fine-grained data sets such as the Wildfish~\cite{zhuang2018wildfish} and iNaturalist~\cite{Horn_2018_CVPR}, the data distributions are usually imbalanced and follow the long-tail distribution, and in some of the categories, the well-labeled training samples are limited, \textit{e.g.,} it is hard to collect large-scale samples of endangered species. How to tackle the fine-grained image classification with limited training data remains an open problem. 

\begin{figure}[t]
	\centerline{
		\includegraphics[width=1\linewidth]{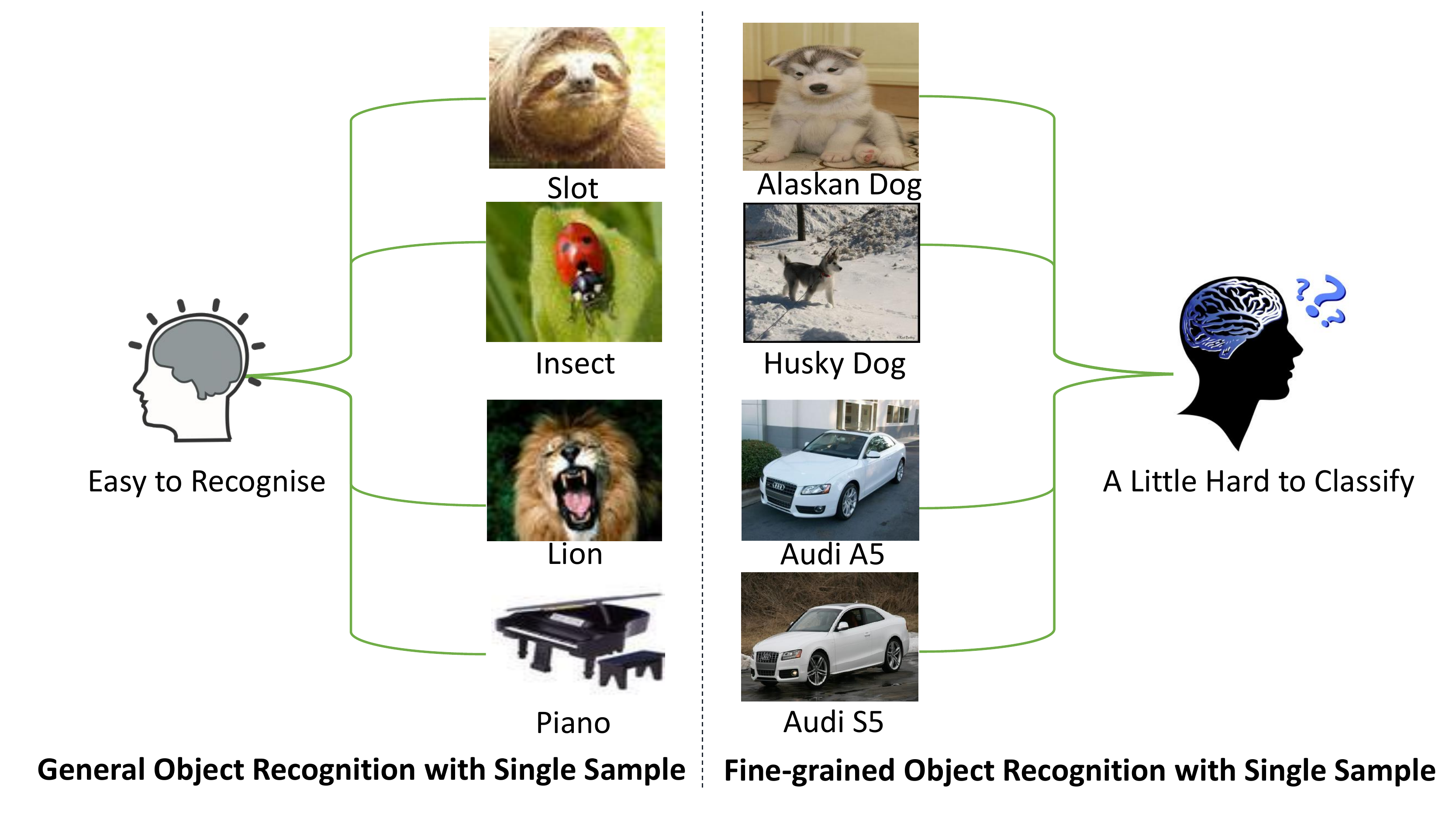}
	}
	\caption{An example of general one-shot learning (Left) and fine-grained one-shot learning (Right). For general one-shot learning, it is easy to learn the concepts of objects with only a single image. However, it is more difficult to distinguish the sub-classes of specific categories with one sample.}
	\label{fig1}
\end{figure} 

Human beings can learn novel generic concepts with only one or a few samples easily. To simulate this intelligent ability, machine few-shot learning is initially identified by Li \textit{et al.}~\cite{fei2006one}. They propose to utilize probabilistic models to represent object categories and update them with a few training examples. Most recently, inspired by the advanced representation learning ability of deep neural networks, deep machine few-shot learning~\cite{vinyals2016matching,snell2017prototypical,Sung_2018_CVPR,liu2018transductive,li2019CovaMNet,li2019DN4} revives and achieves significant improvements against previous methods. 
However, considering the cognitive process of human beings, preschool students can easily distinguish the difference between generic concepts like the `Cat' and `Dog' after seeing a few exemplary images of these animals, but they may be confused about fine-grained dog categories such as the `Husky' and `Alaskan' with limited samples. The undeveloped classification ability of children in processing information compared to adults~\cite{brown1975development,john1986age} indicates that generic few-shot methods cannot cope with the few-shot fine-grained classification task admirably. To this end, in this paper, we focus on dealing with the Few-Shot Fine-Gained (FSFG) classification in a `developed' way. 

\begin{figure*}[t]
	\centerline{
		\includegraphics[width=1\linewidth]{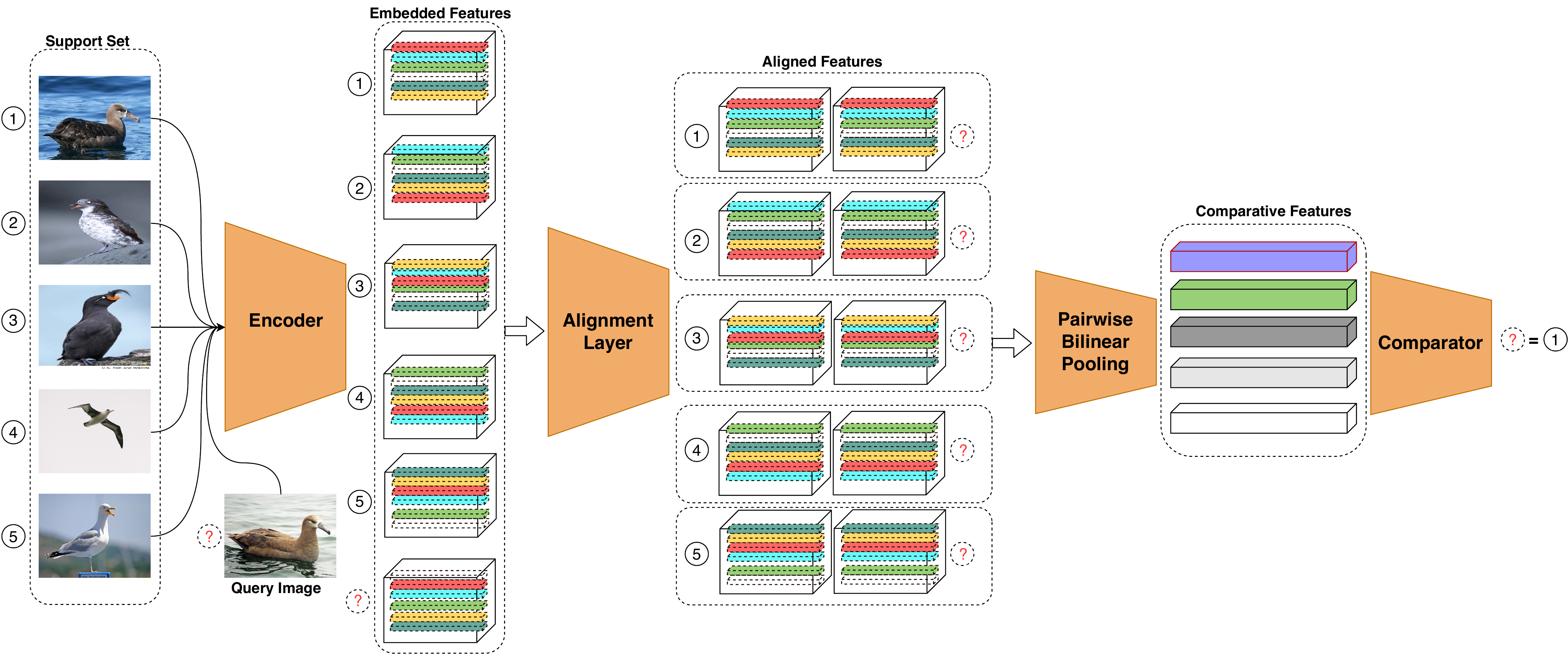}}
	\caption{The framework of LRPABN under the 5-way-1-shot fine-grained image classification setting. The support set contains five labeled samples for each category (marked with numbers) and the query image labeled with a question mark. The LRPPABN model can be divided into four components: Encoder, Alignment Layer, Pairwise Bilinear Pooling, and Comparator. The Encoder extracts coarse features from raw images. Alignment Layer matches the pairs of support and query. Pairwise Bilinear Pooling acts as a fine-grained extractor that captures the subtle features. The Comparator generates the final results. }
	\label{fig2}
\end{figure*}
\begin{figure*}[t]
	\centerline{
		\includegraphics[width = 0.8\textwidth]{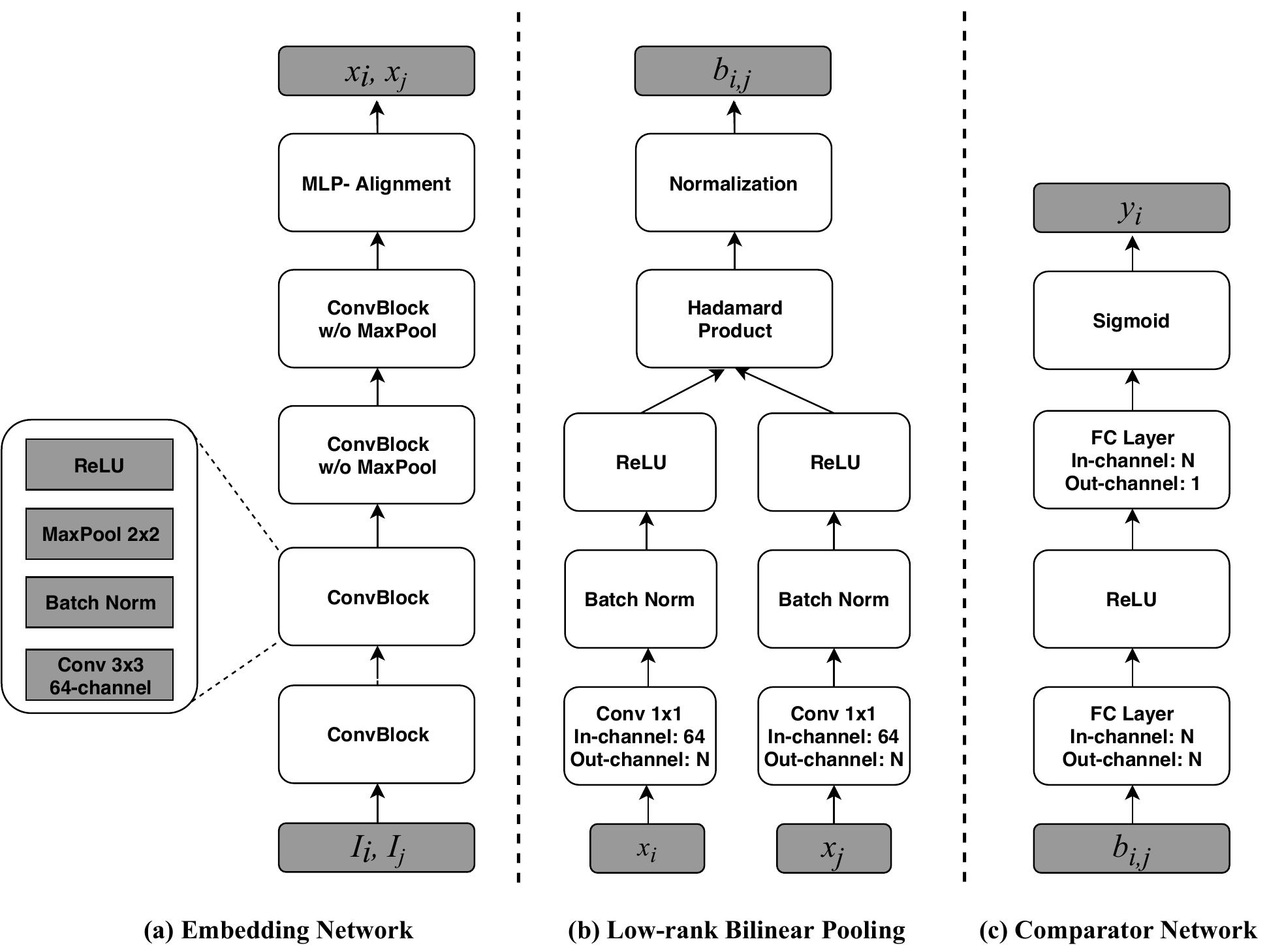}
	}
	\caption{Detailed network architectures used in LRPABN.
(a) The Embedding network with Alignment Layer. (b) Low-Rank Pairwise Bilinear Pooling Layer. (c) The Comparator Network. $I_i$ represents the query image, while $I_j$ is the support image, $x_i, x_j$ are the embedded image features and $b_{i,j}$ represents the comparative bilinear feature. $y_i$ is the predicted label by the comparator.}
	\label{fig3}
\end{figure*}

Wei \textit{et al.}~\cite{wei2018piecewise} recently introduce the FSFG task. Besides establishing the FSFG problem, they propose a deep neural network model named Piece-wise Classifier Mapping (PCM). By adopting the meta-learning strategy on the auxiliary data set, their model can classify different samples in the testing data set with a few labeled samples.
The most critical issue in FSFG is to acquire subtle and informative image features. In PCM, the authors adopt the naive self-bilinear pooling to extract image representations, which widely used in the state-of-the-art fine-grained object classification~\cite{Lin_2015_ICCV,Cui_2017_CVPR,lin2018bilinear}. Then with the operation of bilinear feature grouping, the PCM model can generate low-rank subtle descriptors of the original image. 
Most recently, Li \textit{et al.}~\cite{li2019CovaMNet} propose a covariance pooling~\cite{Li_2018_CVPR} to distillate the image representation of each category. 
These matrix-outer-product based bilinear pooling operations~\cite{li2019CovaMNet,wei2018piecewise} could extract the second-order image features and contains more information than traditional first-order features~\cite{lin2018bilinear}, and thus achieve better performance on FSFG tasks than generic ones.

It is worth noting that both \cite{li2019CovaMNet} and \cite{wei2018piecewise} employ bilinear pooling on the input image itself to enhance the information of original features, which noted as the self-bilinear pooling operation. However, when a human identifies the similar objects, she/he tends to compare them thoroughly in a pairwise way, \textit{e.g.,} comparing the heads of two birds first, then the wings and feet last. Therefore, it is natural to enhance the information during the comparing process when dealing with FSFG classification tasks. Based on this motivation, we propose a novel pairwise bilinear pooling operation on the support and query images to extract the comparative second-order images descriptors for FSFG. 

There are a series of works that address the generic few-shot classification by learning to compare~\cite{vinyals2016matching,snell2017prototypical,Sung_2018_CVPR}, among which the RelationNet~\cite{Sung_2018_CVPR} achieves state-of-the-art performance by combining a feature encoder and a non-linear relation comparator. However, the matching feature extraction in the RelationNet only concatenates the support and query feature maps in depth (channel) dimension and fails to capture nuanced features for the fine-grained classification.
  
To address the above issues, we propose a novel end-to-end FSFG model that captures the fine-grained relations among different classes. This subtle comparative ability of our models is inherently more intelligent than merely modeling the data distribution~\cite{wei2018piecewise,li2019CovaMNet,Sung_2018_CVPR}. The main contributions are summarized as follows:
\begin{itemize}
	\item \textbf{\textit{Pairwise Bilinear Pooling}}. Existing second-order based FSFG methods~\cite{li2019CovaMNet,wei2018piecewise} enhance the individual encoded features by directly applying the self-bilinear pooling operation. However, such an operation fails to capture more nuanced relations between similar objects. Instead, we uncover the fine-grained relations between different support and query image pairs by using matrix outer product operation, which is called pairwise bilinear pooling. Based on the explicit elicitation of correlative information of pair samples, the proposed operation can extract more discriminate features than existing approaches\cite{wei2018piecewise,huang2019compare,Sung_2018_CVPR}. More importantly, we introduce a low-rank approximation for the comparative second-order feature, where a set of co-variance low-rank transform matrices are learned to reduce the complexity of the operation.
	
	\item \textbf{\textit{Effective Feature Alignment}}. The main advantage of self-bilinear based FSFG methods is the enhancement of depth information for individual spatial positions in the image, which is achieved by the matrix outer product operation on convolved feature maps. Inspired by the self-bilinear pooling operation, we design a simple yet effective alignment mechanism to match the pairwise convolved image features. By exploiting the compact image features alignment, the ablation study shows that the proposed alignment mechanism is crucial for the significant improvements against the baseline model, where only the alignment loss is applied~\cite{huang2019compare}.
	
	\item \textbf{\textit{Performance}}. By incorporating the feature alignment mechanism and pairwise bilinear pooling operation, the proposed model achieves the state-of-the-art performances on four benchmark data sets.
\end{itemize}

The preliminary version of the proposed model was published at IEEE ICME-19~\cite{huang2019compare}. The differences between the preliminary version and the new materials are mainly from three aspects:
\begin{itemize}
    \item A more advanced pairwise pooling operation with a low-rank constraint is proposed. Instead of directly operating the matrix-outer-product as the previous version, we propose to learn multiple transformations for fusing the input image features. \textcolor{black}{By applying these transformations, the proposed model generates more compact and discriminative bilinear features than previous ones, which is verified by the coding-pooling theory~\cite{riesenhuber1999hierarchical,aaai2020}. Moreover, we introduce a low-rank approximation of the new bilinear model as our final model to further reduce the computation complexity.}
    \item A novel alignment mechanism is introduced to encourage the input feature pairs of the bilinear operation are matched. \textcolor{black}{Instead of solely relying on the alignment losses, we incorporate a feature position re-arrangement layer with the alignment loss to boost the matching performance.}  
    \item  More comprehensive experimental results analysis and ablation studies are conducted, and the proposed model achieves superior performances against compared models.
\end{itemize}

The rest of this paper is organized as follows: Section \ref{rw} gives a brief introduction of related works on Fine-grained Object classification, Generic Deep Few-shot Learning as well as recent progress in Fine-grained Few-shot Learning. Section \ref{method} presents the proposed LRPABN method, then Section \ref{expt} offers the data sets description, experiment setup, and experimental results analysis. Section \ref{conc} concludes the whole paper in the last.
\section{Related Work}\label{rw}

\subsection{Fine-Grained Object Classification}
Fine-grained object classification has been a trending topic in the computer vision research area for years, and most traditional fine-grained approaches use hand-crafted features as image representations~\cite{xie2014BOG,gao2014learningBOG,zhang2016fusedALIGN}.
However, due to the limited representative capacity of hand-crafted features, the performance of this type of method is moderate.
In recent years, deep neural networks have developed advanced abilities in the feature extraction and function approximation \cite{xu2014multi,he2016deep,gal2016dropout,yao2018discovering,zhang2018multilabel,qiao2018few,zhang2019mangonet}, bringing significant progress in the fine-grained image classification task~\cite{huang2016pbc,xu2017friendMIL,zhang2016PART,zhao2017diversified,huang2016pbc,yao2016coarsePART,peng2018objectPART,zhang2016detectingPART,iscen2015comparisonPART,zhang2014part,Fu_2017_CVPR,Lin_2015_ICCV,Gao_2016_CVPR,Kong_2017_CVPR,Cui_2017_CVPR,Li_2018_CVPR,lin2018bilinear,Suh_2018_ECCV,Yu_2018_ECCV}.

Deep fine-grained classification approaches can be roughly divided into two groups: regional feature-based methods~\cite{huang2016pbc,xu2017friendMIL,zhang2016PART,zhao2017diversified,huang2016pbc,yao2016coarsePART,peng2018objectPART,zhang2016detectingPART,iscen2015comparisonPART,zhang2014part,Fu_2017_CVPR} and global feature-based methods~\cite{Lin_2015_ICCV,Gao_2016_CVPR,Kong_2017_CVPR,Cui_2017_CVPR,Li_2018_CVPR,lin2018bilinear,Suh_2018_ECCV,Yu_2018_ECCV}. 
In fine-grained image classification, the most informative information generally lies in the discriminate parts of the object. Therefore, regional feature-based approaches tend to detect such parts first and then fuse them to form a robustness representation of the object. For instance, Zhang \textit{et al.} \cite{zhang2016PART} firstly combine the R-CNN~\cite{Girshick_2014_CVPR} into the fine-grained classifier with a geometric prior, in which the modified R-CNN generates thousands of proposals, the most discriminate ones are then selected for the object classification. In \cite{peng2018objectPART}, Peng \textit{et al.} adopt two attention modules to localize objects and choose the discriminate parts simultaneously. A spectral clustering method is then employed to align the parts with the same semantic meaning for the prediction.  
However, the classification performance of these models relies heavily on the parts localization. Getting a well-trained part detector needs the input of a large amount of subtle annotated samples, which is infeasible to obtain. Moreover, the sophisticated regional feature fusion mechanism leads to the increasing complexity of the fine-grained classifier.

On the contrary, global feature-based fine-grained methods~\cite{Lin_2015_ICCV,Gao_2016_CVPR,Kong_2017_CVPR,Cui_2017_CVPR,Li_2018_CVPR,lin2018bilinear,Suh_2018_ECCV,Yu_2018_ECCV} extract the feature from the whole image without explicitly localize the object parts.
Bilinear CNN model (BCNN)~\cite{Lin_2015_ICCV} is the first work that adopts matrix outer product operation on the original embedded features to generate a second-order representation for fine-grained classification. Li \textit{et al.}~\cite{Li_2018_CVPR} (iSQRT-COV) further improve the navie bilinear model by using covariance matrices over the last convolutional features as fine-grained features. iSQRT-COV obtains state-of-the-art performance on both generic and fine-grained datasets.

However, the feature dimensions of the second-order models are the square fold of the naive ones, to reduce the computation complexity, Gao \textit{et al.} \cite{Gao_2016_CVPR} propose a compact bilinear pooling operation, which applies Tensor Sketch~\cite{pham2013fast} to reduce the dimensions. Kong \textit{et al.}~\cite{Kong_2017_CVPR} introduce a low-rank co-decomposition of the covariance matrix that fatherly decreases the complexity, while Kim \textit{et al.} \cite{Kim2017} adopt Hadamard product to redefine the bilinear matrix outer product and proposes a factorized low-rank bilinear pooling for multimodal learning. Furthermore, Gao \textit{et al.}~\cite{Yu_2018_ECCV} devise a hierarchical approach for fine-grained classification using a cross-layer factorized bilinear pooling operation. Inspired by the flexibility and effectiveness of the Hadamard product for extracting the second-order features between visual features and textual features in VQA tasks~\cite{Kim2017}, in our LRPABN model, we propose to adopt the factorized bilinear pooling to approximate pairwise second-order statistics for FSFG task. LRPABN achieves better performance compared to the previous models.

\subsection{Generic Deep Few-shot Learning}

The majority of deep few-shot learning methods \cite{pmlr-v70-finn17a,Sachin2017}\cite{vinyals2016matching,snell2017prototypical,Sung_2018_CVPR,liu2018transductive,li2019CovaMNet,li2019DN4} follow the strategy of meta-learning~\cite{phd,Thrun:1998:LL:296635}, which distills the meta-knowledge from batches of auxiliary few-shot tasks. Each auxiliary task mimics the target few-shot tasks with the same support and query images' split. After episodes of training on auxiliary tasks, the trained model can converge speedily to an appreciable local optimum on target data without suffering from the overfitting.

One of the most representative methods is by learning from fine-tuning \cite{chen2019closer},
MAML~\cite{pmlr-v70-finn17a} designs a meta-learning framework that determines the transferable weights for the initialization of the deep neural network. By fine-tuning the network with the limited training samples, the model can achieve reliable performance in a few gradient descent update steps. Moreover, Sachin \textit{et al.} \cite{Sachin2017} propose a gradient-based method that learns well-initialized weights but also an effective LSTM-based optimizer. Different from this type of approach, our model is free from retraining during the meta-testing stage.   

Another class of few-shot learning methods follows the idea of learning to compare~\cite{vinyals2016matching,snell2017prototypical,Sung_2018_CVPR,chen2019closerfewshot,li2019DN4}.
In general, these approaches consist of two main components: a feature embedding network and a similarity metric. These methods aim to optimize the transferable embedding of both auxiliary data and target data. Consequently, the test images can be identified by the simple nearest neighbor classifier~\cite{vinyals2016matching,snell2017prototypical}, deep distance matrix based classifier~\cite{Sung_2018_CVPR}, or cosine-distance based classifier~\cite{chen2019closerfewshot,li2019DN4}. 
Considering the FSFG task requires the more advanced information processing ability, we propose to capture more nuanced features from the images pairs other than the first-order extraction used in leaning to compare approaches.

\subsection{Few-shot Fine-grained  Learning}
Most recently, Wei \textit{et al.} \cite{wei2018piecewise} propose the first FSFG model by employing two sub-networks to tackle the problem jointly. The first component is a self-bilinear encoder, which adopts the matrix outer product operation on convolved features to capture subtle image features, while the second one is a mapping network that learns the decision boundaries of the input data. Li \textit{et al.} \cite{li2019CovaMNet} further replace the naive self-bilinear pooing as the covariance pooling. Moreover, they design a covariance metric to generate relation scores. However, self-bilinear pooling~\cite{wei2018piecewise,li2019CovaMNet} cannot extract comparative features between pairs of images, and the dimension of pooled features is usually large. Pahde \textit{et al.}~\cite{pahde2018cross} propose a cross-modality FSFG model, which embeds the textual annotations and image features into a common latent space. They also introduce a discriminative text-conditional GAN for the sample generation, which selects the representative samples from the auxiliary set. However, it is both computation and time consuming to obtain rich annotations for the fine-grained samples. 
\section{Methodology}\label{method}
In this section, we present the problem formulation of FSFG first. Then the proposed LRPABN model is introduced, including the Low-Rank Pairwise Bilinear Polling operation and Feature Alignment Layer, which are the core parts of LRPABN. The detailed network architecture of LRPABN is given at last.

\subsection{Problem Definition} \label{def}
Given a Fine-Grained target data set $\mathcal{T}:$
\begin{equation}
\begin{split}
&\mathcal{T} = \left\{ \mathcal { B } = \left\{ \left(\overline{x} _ { b } , \overline{y} _ { b } \right) \right\} _ { b = 1 } ^ { K \times \tilde{C} } \right\} \cup \left\{\mathcal { N } = \left\{ \left( \overline{x} _ { v }  \right) \right\} _ { v = 1 } ^ { V } \right\}, \\
&\overline{y} _ { b } \in \{ 1 , \tilde{C} \} , \overline { x } \in   \mathcal { R } ^ { N }, \mathcal{B} \cap \mathcal{N} = \emptyset, V \gg K \times \tilde{C}.
\end{split}
\end{equation} 
For the FSFG task, the target data set $\mathcal{T}$ contains two parts: the labeled subset $\mathcal{ B }$ and the unlabeled subset $\mathcal{ N }$, where samples from each subset are fine-grained images. The model needs to classify the unlabeled data $\overline{x} _ { v }$ from $\mathcal{ N }$ according to a few labeled samples from $\mathcal{ B }$, where $\overline{y} _ { b }$ is the ground-truth label of sample $\overline{x} _ { b }$. If the labeled data in the target data set contains $K$ labeled images for each of $\tilde{C}$ different categories, the problem is noted as $\tilde{C}$-way-$K$-shot.

In order to obtain an ideal model on such a data set, Few-Shot learning usually employs a fully annotated data set, which has similar property or data distribution with $\mathcal{ T }$ as an auxiliary data set $\mathcal{A}$: 
\begin{equation}
\begin{split}
&\mathcal{A} = \left\{ \mathcal { S } = \left\{ \left( x _ { i } , y _ { i } \right) \right\} _ { i = 1 } ^ { L } \right\} \cup \left\{\mathcal { Q } = \left\{ \left( x _ { j } , y _ { j } \right) \right\} _ { j = 1 } ^ { J } \right\}, \\
& y_{i},{y}_{j} \in \{ 1 , {C} \} , x \in   \mathcal { R } ^ { N }, \mathcal{S} \cap \mathcal{Q} = \emptyset, \mathcal{A} \cap \mathcal{T} = \emptyset,
\end{split}
\end{equation}
where  $x_{i}/y_{i}$ and ${x}_{j}/y_{j}$ represent images and their corresponding labels.
In each round of training, the auxiliary data set $\mathcal{A}$ is randomly separated into two parts: support data set $\mathcal{ S }$, and query data set $\mathcal{ Q }$. With setting $L = K \times \tilde{C}$, we can mimic the composition of the target data set in each iteration.
Then $\mathcal{ A}$ is employed to learn a meta-learner $\mathfrak{F}$, which can transfer the knowledge from $\mathcal{A}$ to target data $\mathcal{T}$. Once obtained meta-learner, it can be fine-tuned with labeled target data set $\mathcal{ B }$, and finally, classify the samples from $\mathcal{ N }$ into their corresponding categories \cite{vinyals2016matching,Sung_2018_CVPR,wei2018piecewise,liu2018transductive,huang2019compare,li2019DN4,li2019CovaMNet}. 

\subsection{The proposed LRPABN} \label{LRPABN-1}
The whole framework of LRPABN is shown in Figure~\ref{fig2}, and detailed architecture is given in Figure~\ref{fig3}. 
Different from traditional few-shot embedding structures~\cite{vinyals2016matching,snell2017prototypical,Sung_2018_CVPR}, 
we add the Low-Rank Pairwise Bilinear Pooling to construct the fine-grained image feature extractor. Moreover, we modify the non-linear comparator~\cite{Sung_2018_CVPR} and apply it to the fine-grained task. 
As the Figure~\ref{fig2} shows, given the support set consisting of five classes with one image per class, an Encoder that is trained with the auxiliary data $\mathcal{A}$ can extract the first-order image features from the raw images, then the Alignment Layer coordinates the embedded feature in support set with the query image feature in pairs. Next, the Low-Rank Bilinear Pooling is used to generate the comparative second-order image representation from the embedded feature pairs. Finally, the Comparator assigns the optimal label to the query from support labels in consonance with the similarity between the query and different support classes.  

Pairwise bilinear pooling layer aims to capture the nuanced comparative features of image pairs by employing the bilinear pooling operation, which plays a crucial role in determining the relations between support and query pairs. However, it is natural that if a couple of inputs are not well-matched, the pooled features cannot result in the maximum classification performance gain. Therefore, we introduce an alignment layer which consists of a Multi-Layer Perceptron (MLP) and feature alignment losses to guarantee the registration of the pairs.

\subsubsection{Pairwise Bilinear Pooling Layer}
The Bilinear CNN~\cite{Lin_2015_ICCV} for the image classification can be defined as a quadruple:
\begin{equation}
\begin{split}
&\textit{B-CNNs} = (\mathfrak{E}_{\Rnum{1}}, \mathfrak{E}_{\Rnum{2}}, \mathfrak{f}_{b}, \mathcal{C}),  \\
&\mathfrak{E}: \mathcal{I} \longrightarrow \mathcal{X} \in \mathcal{R}^{c \times h\times w}, \\
&\mathfrak{f}_{b}(\mathcal{I},\mathfrak{E}_{\Rnum{1}}, \mathfrak{E}_{\Rnum{2}}) = \frac{1}{hw} \sum_{i=1}^{hw} f_{\alpha,i} f_{\beta,i}^{T}, 
\label{eq3}
\end{split} 
\end{equation}
where $\mathfrak{E}_{\Rnum{1}}$ and $\mathfrak{E}_{\Rnum{2}}$ are encoders for each input stream, $\mathfrak{f}_{b}$ is the self-bilinear pooling operation, and $\mathcal{C}$ represents the classifier. $\mathcal{I} \in \mathcal{R}^{H \times W \times C}$ is the input image with $H$ height, $W$ width, and $C$ color channels. Through encoder $\mathfrak{E}$, the input image is transformed into a tensor $\mathcal{M} \in \mathcal{R}^{h \times w \times c}$, which has $c$ feature channels, and $h, w$ indicate the height and width of the embedded feature map. Given two encoders $\mathfrak{E}_{\Rnum{1}}: \mathcal{I} \longrightarrow \mathcal{X}_{\alpha} \in \mathcal{R}^{c_1 \times h\times w} $ and $\mathfrak{E}_{\Rnum{2}}: \mathcal{I} \longrightarrow \mathcal{X}_{\beta} \in \mathcal{R}^{c_2 \times h \times w} $,
$f_{\alpha,i} \in \mathcal{R}^{c_1 \times 1}$  and  $f_{\beta,i} \in \mathcal{R}^{c_2 \times 1}$ denote feature vectors at specific spatial location $i$ in each feature map $\mathcal{X}_{\alpha}$ and $\mathcal{X}_{\beta}$, where $i \in [1,hw]$. The pooled feature is a $c_1 \times c_2$ vector. $\mathcal{C}$ is a fully-connected layer trained with cross-entropy loss.

Different from the conventional self-bilinear pooling operates on pairs of embedded features from the same image, in our \textbf{pairwise} bilinear pooling layer, the input pair is generated from the different source sets, \textit{i.e.}, $\mathcal{ I_{A} } \in \mathcal{S}$ and $\mathcal{ I_{B} } \in \mathcal{Q}$. With the encoder $\mathfrak{\tilde{E}}$, the pairwise bilinear pooling $\mathfrak{f}_{pb}$ can be defined as:
\begin{equation}
\begin{split}
&\mathfrak{f}_{pb}(\mathcal{I_{A}},\mathcal{I_{B}}, \mathfrak{\tilde{E}}) = \mathfrak{\tilde{E}}(\mathcal{ I_{A} })\mathfrak{\tilde{E}}(\mathcal{ I_{B} })^{T},\\
&\mathfrak{\tilde{E}}: \mathcal{I} \longrightarrow \mathcal{X} \in \mathcal{R}^{c \times hw}.
\label{eq4}
\end{split} 
\end{equation}
It is worth noting that in the pairwise bilinear pooling, we only have one shared embedding function $\mathfrak{\tilde{E}}$. Different from the self-bilinear pooling that operates on the same input image, pairwise bilinear pooling uses a matrix-outer-product on two different samples.
Equation~(\ref{eq4}) is the pairwise bilinear pooling used in our previous work~\cite{huang2019compare}. 
\textcolor{black}{However, the pooled pairwise feature is a $c_1 \times c_2$ vector, which results in a square growth of the original feature dimension. For example, with an embedding network AlexNet~\cite{krizhevsky2012imagenet}, $c_1 = c_2 = 512$, the pairwise bilinear pooling generates a 512 $\times$ 512 = 262,144-d representation. As reported in~\cite{Gao_2016_CVPR}, in such a high-dimensional feature space, less than 5\% of dimensions are informative. Moreover, recent research~\cite{aaai2020} also indicates that the matrix-outer-product based bilinear pooling suffers from redundancy and burstiness issues because of the rank-one property of bilinear features. The dimensionality of matrix-outer-product based bilinear features incites the heavy computational loads as well as burstiness phenomenons.} 

To overcome this shortcoming of previous proposed pairwise bilinear pooling, inspired by the Factorized Bilinear Pooling~\cite{Kim2017} applied in the visual-question-answer task, we further propose a \textbf{low-rank} pairwise bilinear pooling operation.
For the given $\mathcal{X_{A}}=\left[\mathbf{x}_{1}^{A}, \mathbf{x}_{2}^{A}, \cdots, \mathbf{x}_{hw}^{A}\right]$ and $\mathcal{X_{B}}=\left[\mathbf{x}_{1}^{B}, \mathbf{x}_{2}^{B}, \cdots, \mathbf{x}_{hw}^{B}\right]$ from Equation (\ref{eq4}), where $\mathbf{x}_{j} \in \mathcal{R}^{c \times 1}$ stands for any spatial feature vector in $\mathcal{X}$, $j\in[1, hw]$. The low-rank pairwise bilinear can be formulated as:
\begin{equation}
z_{j}=\left ( \mathbf{x}^A_j \right )^{T} W_{i} \mathbf{x}^B_j,
\label{eq41}
\end{equation}
where $W_{i} \in \mathcal{R}^{c \times c}$ is a projection matrix, $\mathbf{x}^A_j$ and $\mathbf{x}^B_j$ are the feature vectors from $\mathcal{X_{A}}$ and $\mathcal{X_{B}}$ in the same position $j$, separately. Equation~(\ref{eq41}) fuses these feature vectors into a common scalar $z_{j}$. Given a set of projection matrices $\mathcal{W}=\left[W_{1}, W_{2}, \cdots, W_{n}\right] \in \mathcal{R}^{c \times c \times n}$, the redefined bilinear feature of any position $j$ is $\mathbf{z}_{j} = \left[{z}_{1}, z_{2}, \cdots, z_{n}\right]^{T}$.
$n$ is the dimension of this bilinear feature. Then the comparative bilinear representation for the original pairs can be represented as $\mathcal{Z}=\left[\mathbf{z}_{1}, \mathbf{z}_{2}, \cdots, \mathbf{z}_{hw}\right]$. It is worth noticing that Equation (\ref{eq4}) is different from Equation (\ref{eq41}), which adopts projection matrix $W_i$ in learning the bilinear feature. Moreover, in Equation (\ref{eq41}), the dimension of comparative bilinear feature is $n$ that can be far smaller than $c \times c$ in Equation (\ref{eq4}). In this way, the model gets a low-rank approximation for the original comparative bilinear feature.

\textcolor{black}{In Equation~(\ref{eq41}), the learned projection $\mathcal{W}$ requires $c \times c \times n$ parameters, where c = 64 and n = 512 in our implementation, \textit{i.e.,} 2,097,152 parameters in total, which requires a large amount of memory footprint, inference time, and computational complexity.} To solve this problem, we present a low-rank approximation of $W_i$:
\begin{equation}
\begin{split}
    &z_{j}={\left ( ^{}{\mathbf{x}^{A}_j} \right )}^{T} W_{i} \mathbf{x}^B_j \\
    &~~={\left(^{} \mathbf{x}^A_j\right)}^{T} U_{i} V_{i}^{T} \mathbf{x}^B_j \\
    &~~=U_{i}^{T} \mathbf{x}^A_j \circ V_{i}^{T} \mathbf{x}^B_j,
\end{split}
\label{eq42}
\end{equation}
where $U_{i} \in \mathcal{R}^{c \times 1}$ and $V_{i} \in \mathcal{R}^{c \times 1}$, $\circ$ denotes the Hadamard product. Equation (\ref{eq42}) is the \textbf{final form} of low-rank pairwise bilinear pooling, which applies projection matrix and matrix factorization to approximate a full low-rank bilinear model~(Equation (\ref{eq41})). \textcolor{black}{In Equation (6), it needs $2nc$ parameters to generate the pairwise bilinear feature. Therefore, the spatial complexity of the required parameters
is reduced from $\mathcal{O}$$(nc^2)$ to $\mathcal{O}$$(nc)$. It is worth noting that there are two low-rank approximations applied in the final form of our newly proposed model LRPABN. One is to tackle the information redundancy and burstiness issue of the matrix-outer-product based bilinear pooling (Equation (\ref{eq4}) to (\ref{eq41})), the other is to apply the low-rank matrix factorization to approximate the learned transformations (Equation (\ref{eq41}) to (\ref{eq42})).}
The proposed LRPABN is different from \cite{Kim2017,Yu_2018_ECCV}, where \cite{Kim2017} adopts the factorized bilinear pooling to fuse the multi-modal features, and \cite{Yu_2018_ECCV} operates on convolutional features of the same image. Our method conducts on pairs of support and query images. To our best knowledge, LRPABN is the first work that extracts the low-rank bilinear feature from pairs of distinct images for FSFG tasks.

\textcolor{black}{Theoretically, the previous proposed model~\cite{huang2019compare} belongs to the category of matrix-outer-product bilinear pooling, which has been proved as a similarity-based coding-pooling~\cite{riesenhuber1999hierarchical,aaai2020}. As \cite{aaai2020} (Corollary 2) indicates that such bilinear pooling has the unstable dictionary, which is determined by the input pairs, therefore it is inconsistent for all data. This local dictionary can not capture the global geometry of the whole data space, which results in burstiness issues. However, the newly proposed low-rank pairwise bilinear model~(\ref{eq42}) is a type of factorized bilinear coding (Equation (24) in \cite{aaai2020}), which learns a global dictionary from the entire data space in a scalable way, thus achieves better performance than the previous one.}

\subsubsection{Feature Alignment Layer} \label{FAL}
The self-bilinear pooling operates on the same image, which means in any spatial location of the embedded feature pairs, the operating features are entirely aligned. However, since the proposed pairwise bilinear pooling operates on different inputs, the encoded features may not always be matched.
To overcome this obstacle, \textcolor{black}{in our previous work~\cite{huang2019compare}, we devise two alignment losses to match the input pairs in the embedding space simultaneously during the training stage, which aims at encouraging the embedding network to generate well-matched features in the testing stage. However, it may be hard to obtain the desired embedding network that fully aligns feature pairs by merely adopting the alignment losses.}

Therefore, we design a new feature alignment mechanism inspired by the PointNet~\cite{Qi_2017_CVPR}. Given a position transform function $\mathbf{T}$ and the encoded feature $\mathcal{X}=\left[\mathbf{x}_{1}, \mathbf{x}_{2}, \cdots, \mathbf{x}_{hw}\right]$, the transformed feature can be computed as follows:
\begin{equation}
\begin{split}
    &\mathcal{X'} = \mathcal{X}\mathbf{T}, \\
    &s.t.~ \mathbf{T}\mathbf{T}^{T} = \mathbf{I},
\end{split}
\label{regu}
\end{equation}
where $\mathbf{T} \in \mathcal{R}^{hw \times hw}$ and $\mathbf{I}$ is an identity matrix. The transformed feature is noted as $\mathcal{X'}=\left[\mathbf{x'}_{1}, \mathbf{x'}_{2}, \cdots, \mathbf{x'}_{hw}\right]$, in which only the positions of the original feature vectors are rearranged. 
The transform matrix can be learned with a shallow neural network. 
Moreover, to ensure the effectiveness of the alignment, we further design two feature alignment losses as follows:
\begin{equation}
\begin{split}
&\textit{Align}_{loss_{1}}(\mathcal{I_{A}},\mathcal{I_{B}}, \mathfrak{\tilde{E}}) = MSE( \mathfrak{\tilde{E}}(\mathcal{ I_{A}}),\mathfrak{\tilde{E}}(\mathcal{ I_{B} })\mathbf{T}), 
\label{eq5}
\end{split} 
\end{equation}
where $\mathfrak{\tilde{E}}$ is the feature encoder. The first $\textit{Align}_{loss_{1}}$ loss is a rough approximation of two embedded image descriptors that minimizing the Euclidean distances of two transformed features. 

\begin{equation}
\begin{split}
&\textit{Align}_{loss_{2}}(\mathcal{I_{A}},\mathcal{I_{B}}, \mathfrak{O}) = MSE( \mathfrak{O}(\mathcal{ I_{A}}),\mathfrak{O}(\mathcal{ I_{B} })\mathbf{T}),\\
&\mathfrak{{O}}(\mathcal{I})= \sum_{1}^{c}{\mathfrak{\tilde{E}}(\mathcal{ I })}, \mathfrak{\tilde{E}}: \mathcal{I} \longrightarrow \mathcal{X} \in \mathcal{R}^{c \times hw}.
\label{eq6}
\end{split} 
\end{equation}

The second ${Align_{loss_{2}}}$ loss is a more concise feature alignment loss. Inspired by the pooling operation, we sum all the embedded features ($\mathcal{X} \in \mathcal{R}^{c \times hw}$) along with the channel dimension ($\mathcal{R}^{c}$) first. And then, we measure the MSE of summed features. By training with the proposed alignment losses, we encourage the model to automatically learn the matching features to generate a better pairwise bilinear feature. \textcolor{black}{It is worth noting that the alignment mechanism utilizes feature position rearrangement matrix $\mathbf{T}$ on one image features ($\mathfrak{\tilde{E}}(\mathcal{ I_{B}})$) to match the target feature ($\mathfrak{\tilde{E}}(\mathcal{ I_{A}})$). $\mathcal{I_{B}}$ can be either the support or query image, and in our implementation, we choose the support image as $\mathcal{I_{B}}$. Under the supervision of alignment losses, the model can generate more compactly matched feature pairs compared to the previous method.}

\subsubsection{Comparator}
As Figure~\ref{fig2} indicates, after passing through the above layers, the pairwise comparative bilinear features are sent to a comparator. This module aims to learn the relations between the query images and support classes. In the one-way-$K$-shot setting, the support classes are represented by a single image, where for $\tilde{C}$-way-$K$-shot
setting, the support classes are computed as the sum value of embedded features of $K$ images in each class, \textit{i.e.}, for each query image, the model generates $\tilde{C}$ comparative bilinear features corresponding to each class. 
For a pair of query image $i$ and support class $j$, the comparative bilinear feature can be represented as $\mathcal{Z}_{i,j}$, where $\mathcal{Z}=\left[\mathbf{z}_{1}, \mathbf{z}_{2}, \cdots, \mathbf{z}_{hw}\right]$, the relation score of $i$ and $j$ is computed as:
\begin{equation}
\begin{split}
    & r_{i, j}=\mathcal{C}(\mathcal{Z}_{i,j}),\\
    & j=1,2, \ldots W; ~~ i=1,2,\ldots,  K\times \tilde{C},
\end{split}
\label{eq43}
\end{equation}
where $\mathcal{C}$ is the comparator, and $r_{i,j}$ is the relation score of query $i$ and class $j$.

\subsubsection{Model Training}

The training loss $\mathcal{L}$ in our bilinear comparator is the MSE loss, which regresses the relation score to the images label similarity. At a certain iteration during the episodic training, there exists $m$ query features and $n$ support class features in total, $\mathcal{L}$ is thus defined as: 
\begin{equation}
\mathcal{L} =  \sum_{i=1}^{m} \sum_{j=1}^{n}\left(r_{i, j}-
\delta\left(y_{i}=y_{j}\right)\right)^{2},
\label{eq45}
\end{equation}
where $\delta\left(y_{i}=y_{j}\right)$ is the indicator, it equals to one when $y_{i}=y_{j}$ and zeroes otherwise. The LRPABN has two optional alignment losses $\textit{Align}_{loss_{1}}$ and $\textit{Align}_{loss_{2}}$. We back-propagate the gradients when the alignment losses are computed immediately. That is, during the training stage, the model will be updated twice in one iteration.
 
\subsection{Network Architecture}
The detailed network architecture is shown in Figure \ref{fig3}. It consists of three parts: Embedding Network, Low-rank Bilinear Pooling Layer, and Comparator Network.

 \textbf{Embdeeding Network}:
For a fair comparison with the state-of-the-art generic few-shot and FSFG approaches, we adopt the same encoder structure in \cite{vinyals2016matching,snell2017prototypical,Sung_2018_CVPR,liu2018transductive,li2019CovaMNet}. It consists of four convolutional blocks, where each block contains a 2D convolutional layer with a $3 \times3$ kernel and 64 filters, a batch normalization layer, and a ReLU layer. Moreover, for the first two convolutional blocks, a $2\times2$ max-pooling layer is added. For simplicity, we integrate the feature alignment layer into the embedding network as the first-order feature extractor, indicated in Figure \ref{fig3}.(a). Unlike the alignment mechanism used in \cite{Qi_2017_CVPR,peng2018objectPART}, we devise a simple two layers MLP with the Regulation (\ref{regu}). \textcolor{black}{As our alignment mechanism is inspired by PointNet~\cite{Qi_2017_CVPR}, which originally adopts a deeper network to learn the transformation matrix $\mathbf{T}$. However, in FSFG, we find that a shallow MLP network is more efficient in learning a good transformation $\mathbf{T}$.} Besides, two optional alignment losses (\ref{eq5}), (\ref{eq6}) are applied in the alignment layer to generate the well-matched pairwise features. 

\textbf{Low-rank Bilinear Pooling Layer}: For the Low-Rank Pairwise Bilinear Pooling layer in Figure \ref{fig3}.(b), we use a convolutional layer with $1\times1$ kernel followed by the batch normalization and a ReLU layer. The Hadamard product and normalization layers are appended to generate the comparative bilinear features.

\textbf{Comparator Network}: The comparator is made up of two Fully Connected (FC) layers. A ReLU, as well as a Sigmoid nonlinearity layer, are applied to generate the final relation scores, as Figure \ref{fig3}.(c) shows. 
\section{Experiment}\label{expt}
In this section, we evaluate the proposed LRPABN on four widely used fine-grained data sets. First, we give a brief introduction to these data sets. Then we describe the experimental setup in detail. Finally, we analyze the experimental results of the proposed models and compare them with other few-shot learning approaches. For a fair comparison, we conduct two groups of experiments on these data sets, for the first group, we follow the setting, which Wei \textit{et al.}~\cite{wei2018piecewise} and \cite{huang2019compare} used, while for the second group, we follow the newest settings in the recent few-shot methods~\cite{li2019CovaMNet,li2019DN4}.
\subsection{Datasets}
There are four data sets used to investigate the proposed models:

\begin{itemize}
	\item CUB Birds~\cite{WahCUB_200_2011} contains 200 categories of birds and a total of 11,788 images.
	\item DOGS~\cite{KhoslaYaoJayadevaprakashFeiFei_FGVC2011} contains 120 categories of dogs and a total of 20,580 images.
	\item CARS~\cite{KrauseStarkDengFei-Fei_3DRR2013} contains 196 categories of cars and a total of 16,185 images.
	\item NABirds~\cite{Horn_2015_CVPR} contains 555 categories of north American birds and a total of 48,562 images.
\end{itemize}
\begin{table}[t]
	\begin{center}
		\fontsize{9.5}{14}\selectfont
		\caption{The class split of four fine-grained data sets, which is the same as PCM~\cite{wei2018piecewise}. $C_{total}$ is the original number of categories in the data sets, $C_{\mathcal{ A}}$ is the number of categories in separated auxiliary data sets and $C_{\mathcal{ T }}$ is the number of categories in target data sets.} \label{tab:cap}
		
		\begin{tabular}{ | c | c | c | c | c| } \hline
			\text { data set } & { \text{CUB Birds}  } & { \text { DOGS } } & { \text { CARS } } & {\text{NABirds}} \\ 
			\hline \hline
			$C _ { \text { total } }$ & { 200 } & { 120 } & { 196 } &{555} \\ 
			$C _ { \mathcal{ A}  }$ & { 150 } & { 90 } & { 147 }&{416} \\ 
			$C _ { \mathcal { T } }$ & { 50 } & { 30 } & { 49 }&{139} \\ \hline 
		\end{tabular} 
	\end{center}
\end{table}
In Section~\ref{def}, we randomly divide these data sets into two disjoint sub-sets: the auxiliary data set $\mathcal{ A}$, and the target data set $\mathcal{ T }$. For the first group of experiments, we use the splits of PCM~\cite{wei2018piecewise}, as shown in Table~\ref{tab:cap}. For the second group, we adopt the data set splits of Li's~\cite{li2019CovaMNet,li2019DN4}, as indicated in Table~\ref{tab:cap2}. Both of these methods do not use the NABirds data set. Thus, for this data set only, we do our splits.

\subsection{Experimental Setup}\label{setup}
In each round of training and testing, for the one-shot image classification setting, the support sample number in each class equals 1 (in both $\mathcal{ B }$ and $\mathcal{ S }$, $K = 1$). Therefore, we use the embedded features of these support samples as the class features, \textit{i.e.}, $\mathfrak{\tilde{E}}(\mathcal{ I_{B} })$. For the few-shot setting, we extract the class features by summing all the embedded support features in each category. 
In our experiments, we compare the below FS as well as FSFG approaches:

\textbf{The state-of-the-art methods:}
\begin{itemize}
    \item RelationNet \cite{Sung_2018_CVPR}, a state-of-the-art generic few-shot method proposed in CVPR 2018. It uses a mini-network to learn the similarity between the query image and the support class.
    \item DN4~\cite{li2019DN4}, the newest generic few-shot method published in CVPR 2019. By using a deep nearest neighbor neural network, DN4 can aggregate the discriminative information of local features and thus improve the final classification performance.
    \item PCM~\cite{wei2018piecewise}, the first FSFG model published in IEEE TIP 2019. It adopts a self-bilinear model to extracts the fine-grained features of the image and achieves excellent performance on several FSFG tasks.
    \item CovaMNet~\cite{li2019CovaMNet}, a newest FSFG model published in AAAI 2019. It replaces the bilinear pooling with covariance bilinear pooling and achieves state-of-the-art performance on FSFG classification.
\end{itemize}
 
\textbf{The PABN models}~\cite{huang2019compare}, our previous work for FSFG tasks that uses pairwise bilinear pooling~(\ref{eq4}) without feature alignment transform function~(\ref{regu}): 
\begin{itemize}
    \item PABN$_{w/o}$, this model does not use alignment loss on embedded pair features.
    \item PABN$_{niv}$ and PABN$_{cpt}$ are the models that adopt the alignment loss $Align_{loss_{1}}$ and $Align_{loss_{2}}$ for feature alignment, separately. As Section \ref{FAL} discussed, $Align_{loss_{1}}$ loss is a naive alignment loss where $Align_{loss_{2}}$ is a more compact loss.
\end{itemize}

\textbf{The PABN+ models}, these models apply the proposed alignment layer into PABN models, which aims to investigate the effectiveness of the proposed feature alignment transform function~(\ref{regu}):
\begin{itemize}
    \item PABN+$_{niv}$ and PABN+$_{cpt}$ are the models that adopt the alignment loss $Align_{loss_{1}}$ and $Align_{loss_{2}}$ in the alignment layer ~(\ref{regu}).
    \item PABN+$_{cons}$ adopts Cosine loss on the embedded features in the alignment layer~(\ref{regu}).
\end{itemize}

\textbf{The LRPABN models}, we replace the naive pairwise bilinear pooling~(\ref{eq4}) with the proposed low-rank bilinear pooling~(\ref{eq42}), and apply the proposed novel feature alignment layer~(\ref{regu}) into the LRPABN models:
\begin{itemize}
    \item LRPABN$_{niv}$ and LRPABN$_{cpt}$, which use the alignment loss $Align_{loss_{1}}$ and the loss $Align_{loss_{2}}$ in the alignment layer, respectively.
\end{itemize}

\begin{table}[t]
	\begin{center}
		\fontsize{9.5}{14}\selectfont
		\caption{The class split of four fine-grained data sets which is the same as \cite{li2019CovaMNet,li2019DN4}. $C_{total}$ is the original number of categories in the data sets, $C_{\mathcal{ A}.Train}$ is the number of training data categories in the auxiliary data sets, $C_{\mathcal{ A}.Val}$ is the number of validation data categories in separated auxiliary data sets and $C_{\mathcal{ T }}$ is the number of categories in target data sets.} \label{tab:cap2}
		
		\begin{tabular}{ | c | c | c | c | c| } \hline
			\text { data set } & { \text{CUB Birds}  } & { \text { DOGS } } & { \text { CARS } } & {\text{NABirds}} \\ 
			\hline \hline
			$C _ { \text { total } }$ & { 200 } & { 120 } & { 196 } &{555} \\ 
			$C _ { \mathcal{ A}.Train  }$ & { 120 } & { 70 } & { 130 }&{350} \\ 
			$C _ { \mathcal{ A}.Val  }$ & { 30 } & { 20 } & { 17 }&{66} \\ 
			$C _ { \mathcal { T } }$ & { 50 } & { 30 } & { 49 }&{139} \\ \hline 
		\end{tabular} 
	\end{center}
\end{table}

In the first experiment, the LRPABN models are compared with RelationNet, PCM, and our previous proposed PABN models. We follow the data splits (Table \ref{tab:cap}) of PCM and PABN. All of these approaches do not contain the validation data set.

In the second experiment, besides the RelationNet, PABN+ models, and the proposed LRPABN models, we compare the newest state-of-the-art few-shot method DN4 and the newest FSFG approach CovaMNet. To fair compare, we use the same data splits (Table \ref{tab:cap2}) and the training strategy of DN4 and CovaMNet.

\begin{table*}[t]  
	\centering  
	\fontsize{8.9}{16.5}\selectfont  
	\caption{Few-shot classification accuracy (\%) comparisons on four fine-grained data sets. The second-highest-accuracy methods are highlighted in blue color. The highest-accuracy methods are labeled with the red color. `-' denotes not reported. All results are with $95\%$ confidence intervals where reported.} \label{tab2}
	\begin{tabular}{|c|c|c|c|c|c|c|c|c|}  
		\hline
		\multirow{2}{*}{Methods} & 	\multicolumn{2}{c|}{CUB Birds}&\multicolumn{2}{c|}{CARS }&\multicolumn{2}{c|}{DOGS}&\multicolumn{2}{c|}{NABirds}  \cr\cline{2-9}
		&1-shot & 5-shot & 1-shot & 5-shot&1-shot & 5-shot & 1-shot & 5-shot \\
		\hline  \hline
		PCM~\cite{wei2018piecewise} & 42.10$\pm$1.96 & 62.48$\pm$1.21& 29.63$\pm$2.38 & 52.28$\pm$1.46 & 28.78$\pm$2.33 & 46.92$\pm$2.00  & - & -\\ \hline
		RelationNet &63.77$\pm$1.37 &74.92$\pm$0.69& 56.28$\pm$0.45 & 68.39$\pm$0.21 & 51.95$\pm$0.46 & 64.91$\pm$0.24 & 65.17$\pm$0.47 & 78.35$\pm$0.21 \\ \hline
		PABN$_{w/o}$ & 65.99$\pm$1.35 &76.90$\pm$0.21  &55.65$\pm$0.42   &67.29$\pm$0.23   & 54.77$\pm$0.44 &65.92$\pm$0.23  &{67.23$\pm$0.42 } &{79.25$\pm$0.20 }   \\ \hline
		PABN$_{niv}$ &65.04$\pm$0.44  & 76.46$\pm$0.22  &55.89$\pm$0.42   & 68.53$\pm$0.23  &54.06$\pm$0.45  &{65.93$\pm$0.24 } &66.62$\pm$0.44  &{79.31$\pm$0.22 }  \\ \hline  
		PABN$_{cpt}$ &\textcolor{blue}{\textbf{66.71$\pm$0.43}} & 76.81$\pm$0.21  & 56.80$\pm$0.45  & 68.78$\pm$0.22 &{\textcolor{red}{\textbf{55.47$\pm$0.46 }}} & 66.65$\pm$0.23 &{67.02$\pm$0.43 } &79.02$\pm$0.21   \\ \hline
		\hline
		PABN+$_{niv}$& 66.68$\pm$0.42 & 76.83$\pm$0.22 & 55.35$\pm$0.44 &67.67$\pm$0.22 &54.51$\pm$0.45 &66.60$\pm$0.23 &66.60$\pm$0.44 &\textbf{\textcolor{red}{{81.07$\pm$0.20}}}\\ \hline
		PABN+$_{cpt}$& 65.44$\pm$0.43& 77.19$\pm$0.22 & 57.36$\pm$0.45 & 69.30$\pm$0.22 & 54.66$\pm$0.45 & \textcolor{blue}{\textbf{66.74$\pm$0.22}} & 67.39$\pm$0.43 & 79.95$\pm$0.21 \\ \hline
		PABN+$_{cos}$& 66.45$\pm$0.42 &  {\textcolor{red}{\textbf{78.34$\pm$0.21}}} &57.44$\pm$0.45 &68.59$\pm$0.22 &54.18$\pm$0.44 &65.70$\pm$0.23 &66.74$\pm$0.44 &80.58$\pm$0.20 \\ \hline
		\hline
		LRPABN$_{niv}$& 64.62$\pm$0.43 &\textcolor{blue}{\textbf{78.26+0.22}} &\textcolor{blue}{\textbf{59.57$\pm$0.46}} & {\textcolor{red}{\textbf{74.66$\pm$0.22}}} & \textcolor{blue}{\textbf{54.82$\pm$0.46}} &66.62$\pm$0.23 & {\textcolor{red}{\textbf{68.40$\pm$0.44}}} &80.17$\pm$0.21 \\ \hline
		LRPABN$_{cpt}$& {\textcolor{red}{\textbf{67.97$\pm$0.44}}} &78.04$\pm$0.22 & {\textcolor{red}{\textbf{63.11$\pm$0.46}}} &\textcolor{blue}{\textbf{72.63$\pm$0.22}} &54.52$\pm$0.47 &\textbf{\textcolor{red}{67.12$\pm$0.23}} &\textcolor{blue}{\textbf{68.04$\pm$0.44}} &\textcolor{blue}{\textbf{{80.85$\pm$0.20}}} \\ \hline
	\end{tabular}  
\end{table*} 
\begin{table*}[t]  
	\centering  
	\fontsize{9.2}{16.5}\selectfont  
	\caption{Few-shot classification accuracy (\%) comparisons on four fine-grained data sets. The highest-accuracy and second-highest-accuracy methods are highlighted in red and blue, separately. All results are with $95\%$ confidence intervals where reported.} \label{tab3}
	\begin{tabular}{|c|c|c|c|c|c|c|c|c|}  
		\hline
		\multirow{2}{*}{Methods} & 	\multicolumn{2}{c|}{CUB Birds}&\multicolumn{2}{c|}{CARS }&\multicolumn{2}{c|}{DOGS}&\multicolumn{2}{c|}{NABirds}  \cr\cline{2-9}
		&1-shot & 5-shot & 1-shot & 5-shot&1-shot & 5-shot & 1-shot & 5-shot \\
		\hline  \hline
		RelationNet &59.82$\pm$0.77 &71.83$\pm$0.61 &56.02$\pm$0.74 &66.93$\pm$0.63 &44.75$\pm$0.70 &58.36$\pm$0.66 &64.34$\pm$0.81 & 77.52$\pm$0.60  \\ \hline
		CovaMNet  &58.51$\pm$0.94 &71.15$\pm$0.80 &56.65$\pm$0.86 &71.33$\pm$0.62 & {\textcolor{red}{\textbf{49.10$\pm$0.76}}}& \textcolor{blue}{\textbf{63.04$\pm$0.65}}&60.03$\pm$0.98 & 75.63$\pm$0.79  \\ \hline
		DN4  & 55.60$\pm$0.89 & {\textcolor{red}{\textbf{77.64$\pm$0.68}}} &\textcolor{blue}{\textbf{59.84$\pm$0.80}} & {\textcolor{red}{\textbf{88.65$\pm$0.44}}} &45.41$\pm$0.76 &  {\textcolor{red}{\textbf{63.51$\pm$0.62}}} & 51.81$\pm$0.91 &  {\textcolor{red}{\textbf{83.38$\pm$0.60}}}  \\ \hline
		\hline
		PABN+$_{niv}$ & \textcolor{blue}{\textbf{63.56$\pm$0.79}} & 75.23$\pm$0.59 & 53.39$\pm$0.72 & 66.56$\pm$0.64 & 45.64$\pm$0.74 & 58.97$\pm$0.63 & \textcolor{blue}{\textbf{66.96$\pm$0.81}} & 80.73$\pm$0.57  \\ \hline
		PABN+$_{cpt}$  & 63.36$\pm$0.80 & 74.71$\pm$0.60 & 54.44$\pm$0.71 & 67.36$\pm$0.61 & 45.65$\pm$0.71 & 61.24$\pm$0.62 & 66.94$\pm$0.82 & 79.66$\pm$0.62  \\ \hline
		PABN+$_{cos}$  &62.02$\pm$0.75 & {75.35$\pm$0.58} & 53.62$\pm$0.73 & 67.15$\pm$0.60 & 45.18$\pm$0.68 & 59.48$\pm$0.65 & 66.34$\pm$0.76 & 80.49$\pm$0.59  \\ \hline 
		\hline
		LRPABN$_{niv}$  &62.70$\pm$0.79 & 75.10$\pm$0.61 & 56.31$\pm$0.73 & 70.23$\pm$0.59 & \textcolor{blue}{\textbf{46.17$\pm$0.73}} & 59.11$\pm$0.67 & 66.42$\pm$0.83 & 80.60$\pm$0.59  \\ \hline
		LRPABN$_{cpt}$  &  {\textcolor{red}{\textbf{63.63$\pm$0.77}}} & \textcolor{blue}{\textbf{76.06$\pm$0.58}} &  {\textcolor{red}{\textbf{60.28$\pm$0.76}}} & \textcolor{blue}{\textbf{73.29$\pm$0.58}} & 45.72$\pm$0.75 & 60.94$\pm$0.66 &  {\textcolor{red}{\textbf{67.73$\pm$0.81}}} & \textcolor{blue}{\textbf{81.62$\pm$0.58}}  \\ \hline
	\end{tabular}  
\end{table*}

For all the comparing methods, we conduct both 5-way-1-shot and 5-way-5-shot classification experiments. In the training stage of the first group of experiments, both 5-way-1-shot and 5-way-5-shot experiments have 15 query images, which means there are $15 \times 5 + 1 \times 5 = 80$ images and $15 \times 5 + 5 \times 5 = 100$ images in each mini-batch, respectively. 
For the testing stage, we follow the RelationNet~\cite{Sung_2018_CVPR} that have one query for 5-way-1-shot and five queries for 5-way-5-shot in each mini-batch. In both the training and testing stages of the second group of experiments, we randomly select 15 and 10 queries from each category for the 5-way-1-shot and 5-way-5-shot settings, which is the same setting with \cite{li2019CovaMNet,li2019DN4}. 

\textcolor{black}{For fair comparisons, we select the optimal models using the same validation strategies as \cite{Sung_2018_CVPR} for the first group of experiments and \cite{li2019DN4,li2019CovaMNet} for the second group of experiments, separately.} In the first group, we randomly sample and construct 100,000 episodes to train the LRPABN and PABN+ models. In each episode, there only contains one learning task, while in the second group, we randomly select 10,000 episodes for training, and in each episode, 100 tasks are randomly batched to train the models. For LRPABN models, we set the dimension of the pairwise bilinear feature as 512, where the feature dimension of PABN and PABN+ is $64 \times 64 = 4096$. \textcolor{black}{In training, the learning rate of parameters is decayed by 0.5 every 10,000 epochs using the StepLR schedule in PyTorch.} We resize all the input images from all data sets to $84 \times 84$. All experiments use Adam optimize method with an initial learning rate of 0.001, and all models are trained end-to-end from scratch.


\begin{table*}[t]  
	\centering  
	\fontsize{9.8}{15.0}\selectfont  
	\caption{Ablation study of LRPABN with different components. The results are reported with 95\% confidence intervals. Model size indicates the number of parameters for each model, and the Inference Time is the testing time for each input query image.} \label{ABS}
	\begin{tabular}{|c|c|c|c|c|c|}  
		\hline
		\multirow{2}{*}{Methods} & 	\multicolumn{5}{c|}{CUB data set} \cr\cline{2-6}
		&1-shot (\%) & 5-shot (\%) & Model Size & Inference Time ($10^{-3}$ s) & Bilinear Feature Dim \\
		\hline
		PABN$_{cpt}$~\cite{huang2019compare}  & 66.71$\pm$0.43 & 76.81$\pm$0.21 & 375,361 & 8.65& 4096 \\ 
		PABN+$_{cpt}$  & 65.44$\pm$0.43  & 77.19$\pm$0.22 & 505,682 & 8.94 & 4096 \\ 
		PABN$_{new}$  & 67.39$\pm$0.43 & 78.87$\pm$0.21  & 2,373,819 & 78.40 & 512\\ 
		LRPABN  & 66.56$\pm$0.43  & 77.60$\pm$0.22 & 213,930 & 2.23 & 512 \\
		LRPABN$_{only\_cpt}$  & 66.72$\pm$0.44 & 77.98$\pm$0.21 & 213,930 & 2.23 & 512 \\ 
	    LRPABN$_{cpt}$  & 67.97$\pm$0.44  & 78.04$\pm$0.22 & 344,251 & 2.53 & 512 \\
	    DN4~\cite{li2019DN4} & 60.02$\pm$0.85 & 79.64$\pm$0.67 & 112,832 & 15.20 & - \\
	   \hline
	\end{tabular}  
\end{table*}

\begin{table}[t]  
	\centering  
	\fontsize{8.7}{16.5}\selectfont  
	\caption{Discussion about input image size for FSFG.} \label{size}
	\begin{tabular}{|c|c|c|c|}  
		\hline
		\multirow{3}{*}{Methods} & 	\multicolumn{3}{c|}{CUB data set} \cr\cline{2-4}
		&1-shot (\%) & 5-shot (\%) & Image Size \\
		\hline
		PCM:AlexNet~\cite{wei2018piecewise}  & 42.10$\pm$1.96 & 62.48$\pm$1.21 & 224 $\times$ 224 \\ 
	    LRPABN$_{cpt}$:AlexNet  & 59.34$\pm$0.48 & 69.08$\pm$0.24 & 224 $\times$ 224\\ 
		LRPABN$_{cpt}$:AlexNet  & 66.19$\pm$0.46 & 75.05$\pm$0.23 & 448 $\times$ 448\\
		LRPABN$_{cpt}$:Conv4  & 67.97$\pm$0.44  & 78.04$\pm$0.22  & 84 $\times$ 84 \\
		DN4:Conv4~\cite{li2019DN4} & 60.02$\pm$0.85 & 79.64$\pm$0.67 & 84 $\times$ 84 \\
	   \hline
	\end{tabular}  
\end{table}

\begin{figure*}[t]
	\centering
	\subfloat[By LRPABN-512] {\includegraphics[width=0.5\textwidth]{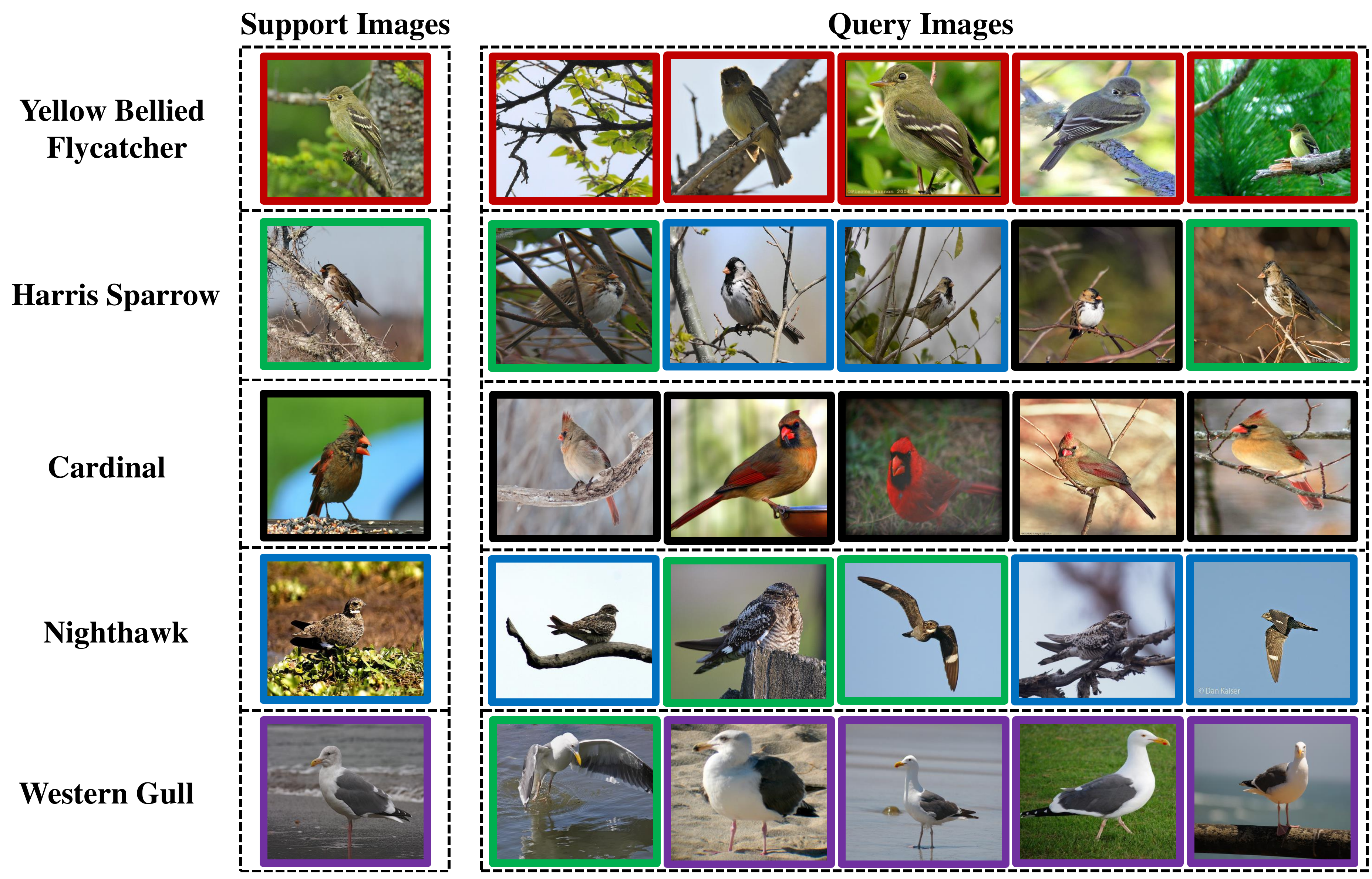} \label{sub001}} 
	\subfloat[By LRPABN-128] {\includegraphics[width=0.5\textwidth]{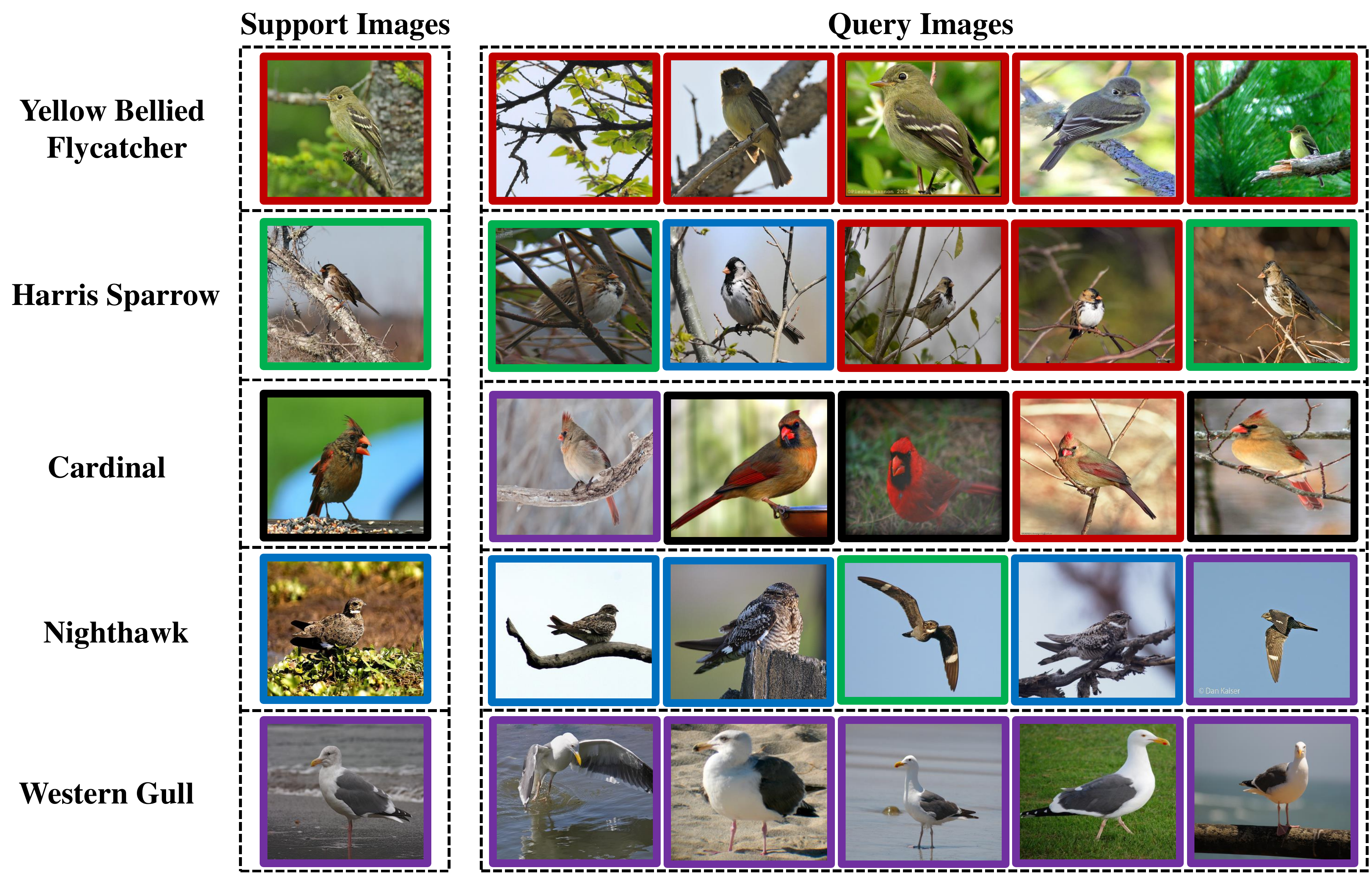} \label{sub002}}\quad
	\subfloat[By PABN+] {\includegraphics[width=0.5\textwidth]{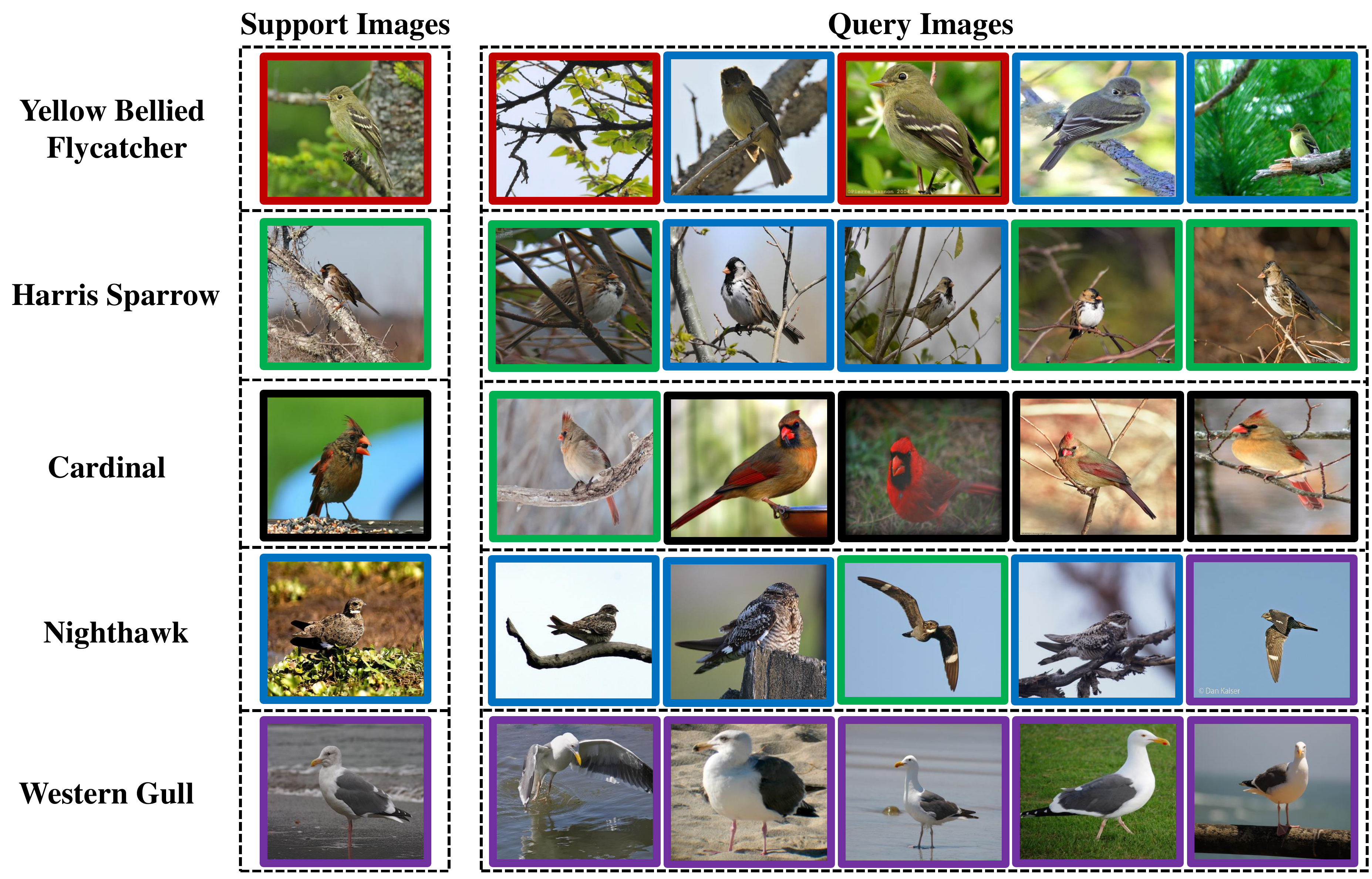} \label{sub003}}
	\subfloat[By RelationNet] {\includegraphics[width=0.5\textwidth]{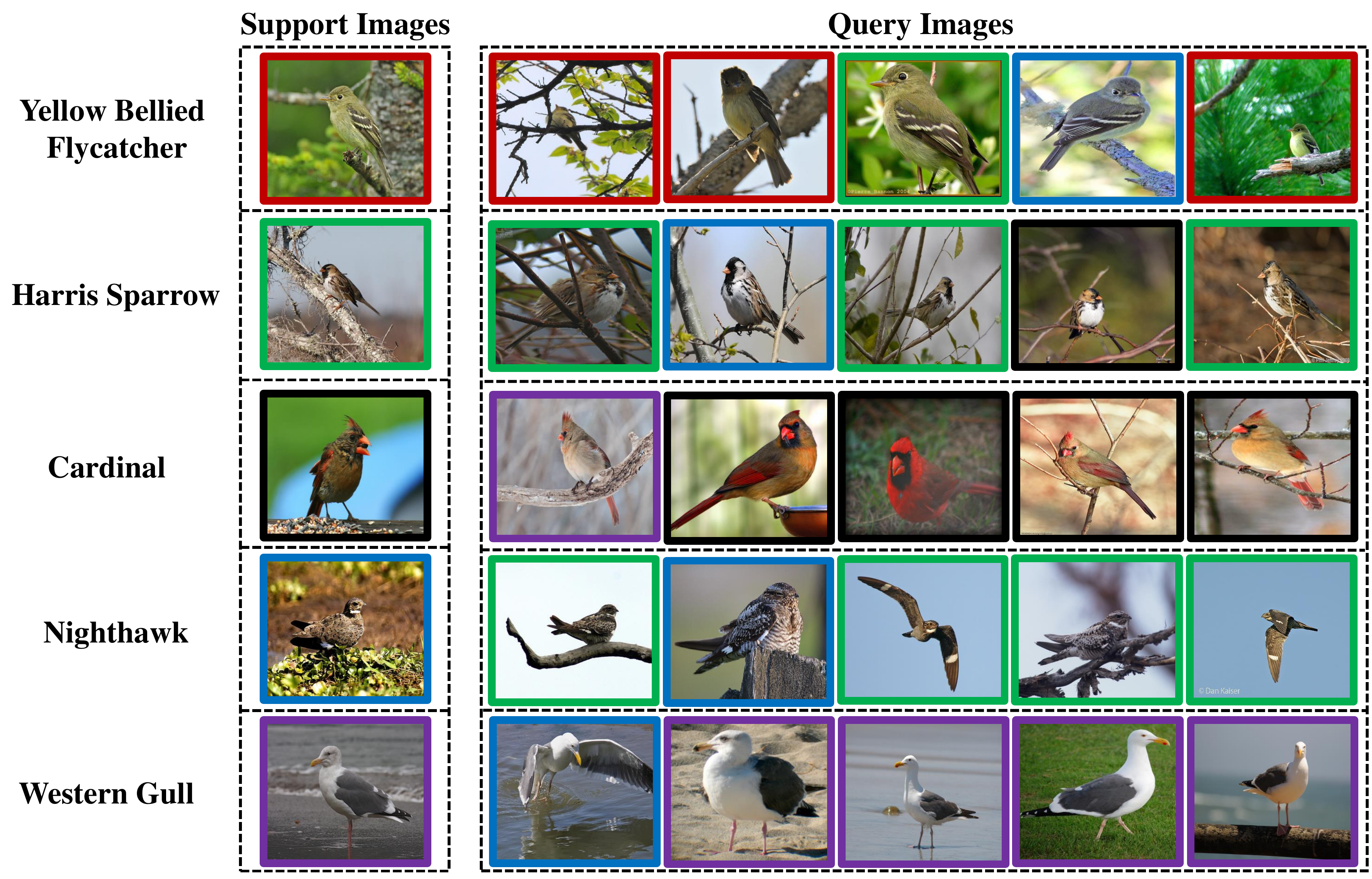} \label{sub004}}
	\caption{Some visual classification results of comparing methods over CUB Birds data set. All the approaches use the same data batch under the five-way-one-shot setting, and for each class, we randomly select five query images as the testing data. We adopt five colors to label the support classes separately. As to the query images, we label the images with the color corresponding to the class label predicted by different models. }
	\label{Imshow}
\end{figure*}
\subsection{Results and Analysis}\label{Res}
To the best of our knowledge, there are only a few methods proposed for FSFG image classification~\cite{wei2018piecewise,huang2019compare,li2019CovaMNet,pahde2018cross,yao2017one}. \cite{pahde2018cross} uses larger auxiliary data set than our methods, and \cite{yao2017one} is only applied for image retrieval tasks. It is unfair to compare these methods directly. Therefore we compare our LRPABN with PCM \cite{wei2018piecewise}, PABN~\cite{huang2019compare}, and CovaMNet~\cite{li2019CovaMNet}. We also compare our methods with the state-of-the-art generic few-shot learning method RelationNet~\cite{Sung_2018_CVPR} and DN4~\cite{li2019DN4}. The original implementation of RelationNet does not report the results on four fine-grained data sets. For fair comparisons, we use the open-source code of the RelationNet\footnote{\url{https://github.com/floodsung/LearningToCompare_FSL}} to conduct the FSFG image classification on these data sets.

In the first group of experiments, we compute both one-shot and five-shot classification accuracies on the four data sets by averaging on 10,000 episodes in testing. We show the experimental results of 10 compared models in Table \ref{tab2}. As the table shows, the proposed LRPABN models achieve significant improvements on both 1-shot and 5-shot classification tasks on all data sets compared to the state-of-the-art FSFG methods and generic few-shot methods, which indicates the effectiveness of the proposed framework. 

More specifically, the LRPABN, PABN+, and PABN models~\cite{huang2019compare} both obtain around 10\% to 30\% higher in classification accuracy than PCM~\cite{wei2018piecewise}, which demonstrates that the comparative pairwise bilinear feature outperforms the self-bilinear feature on FSFG tasks. Besides, the pairwise bilinear feature-based approaches achieve better classification performances than RelationNet~\cite{Sung_2018_CVPR} that validates the extraction of second-order image descriptors surpasses the naive concatenation of feature pairs~\cite{Sung_2018_CVPR} for FSFG problems.

From Table~\ref{tab2}, compared to PABN models, PABN+ and LRPABN models obtain a definite classification performance boost. For instance, the PABN+$_{niv}$ gains $1.64\%$ and $0.37\%$ improvements over PABN$_{niv}$ in one-shot and five-shot setting on CUB Birds data, while LRPABN$_{cpt}$ achieves $1.26\%$ and $1.23\%$ improvements over PABN$_{cpt}$ in one-shot and five-shot setting on CUB Birds data set.
These results demonstrate that the effectiveness of the proposed feature alignment layer. 
It can be observed from Table \ref{tab2} that LRPABN models achieve the best or second-best classification performance on nearly all data sets compared to other methods under various experimental settings. 
For CARS data, the LRPABN$_{cpt}$ obtains $5.67\%$, $6.31\%$, $6.83\%$ significant improvements over PABN+$_{cos}$, PABN$_{cpt}$ and RelationNet on 1-shot-5-way task, while achieves $5.36\%$, $5.88\%$, $6.27\%$ improvements against PABN+$_{cpt}$, PABN$_{cpt}$ and RelationNet on 5-shot-5-way setting, which validates the effectiveness of our low-rank pairwise bilinear pooling. It is worth noting that the dimension of the pairwise bilinear feature in LRPABN is 512, where the corresponding feature dimension of PABN and PABN+ is 4096. 
LRPABN models adopt the low-rank factorized bilinear pooling operation, which can learn a set projection transform functions fusing the feature pairs, as discussed in Equation~(\ref{eq42}). Each of the projection function represents a pattern of coalescing the image pairs in feature channels over all the matching positions. Meanwhile, the naive pairwise bilinear pooling in the PABN and PABN+ approaches only applies the matrix outer product on feature pairs once to merge them. Therefore, the LRPABN models can obtain more types of feature extraction than PABN and PABN+ models, which in turn achieves better performance with smaller feature dimensions.

For a further analysis of our models, we conduct an additional experiment on these four data sets comparing the LRPABN models with DN4 and  CovaMNet. In this experiment, we also compare the PABN+ models. Moreover, we use the same setting to rerun the RelationNet on four data sets as the baseline method. We follow the same data set split with DN4 and CovaMNet, the original papers of these two papers do not report the results on CUB Birds (CUB-2011)~\cite{WahCUB_200_2011} and NABirds~\cite{Horn_2015_CVPR}, so we use the open released codes of DN4\footnote{\url{https://github.com/WenbinLee/DN4}} and CovaMNet\footnote{\url{https://github.com/WenbinLee/CovaMNet}} to get the results. During the test, 600 episodes are randomly selected from the data.
\begin{figure*}[t]
	\centering
	\subfloat[] {\includegraphics[width=0.49\textwidth]{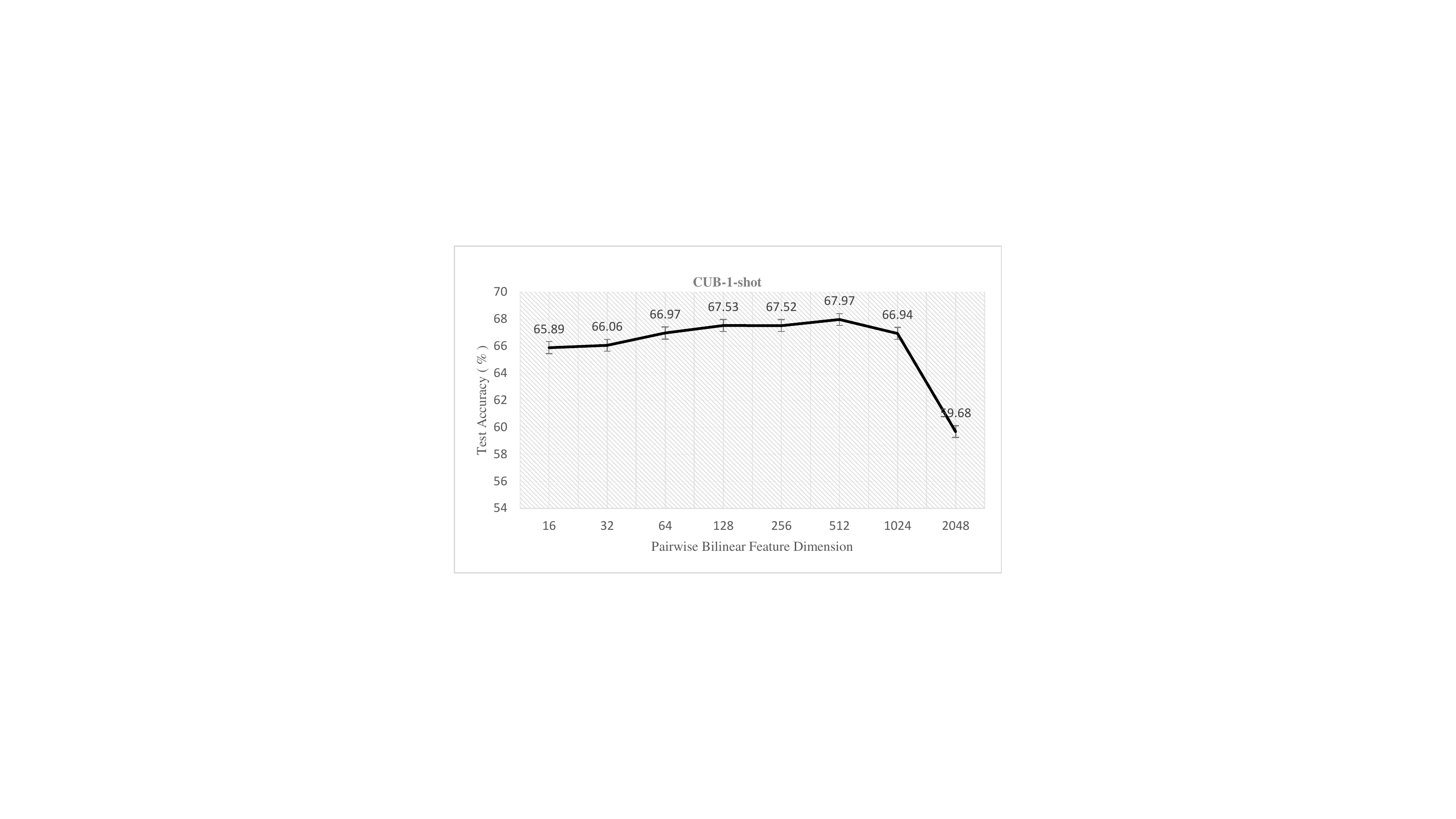} \label{sub1}}
	\subfloat[] {\includegraphics[width=0.49\textwidth]{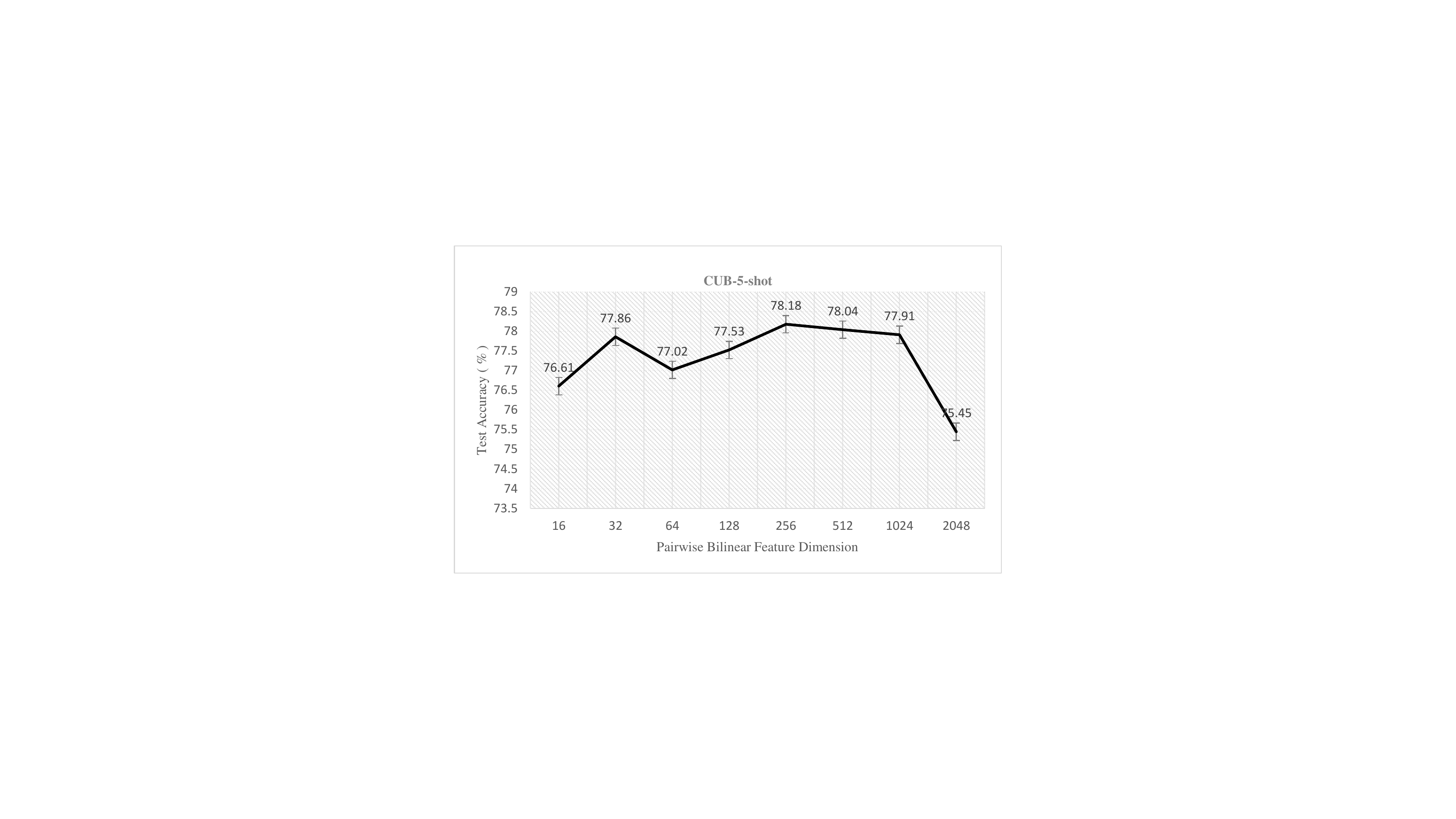}
	\label{sub2}}
	\caption{The pairwise bilinear feature dimension selection experiment. In each sub-figure, the horizontal axis denotes the dimension of the pairwise bilinear feature and the vertical axis represents the test accuracy rate. \ref{sub1} is the one-shot experiment and \ref{sub2} is the five-shot experiment on CUB data set.}
	\label{figdim}
\end{figure*}
\begin{figure*}[t]
	\centering
	\subfloat[By RelationNet] {\includegraphics[width=0.245\textwidth]{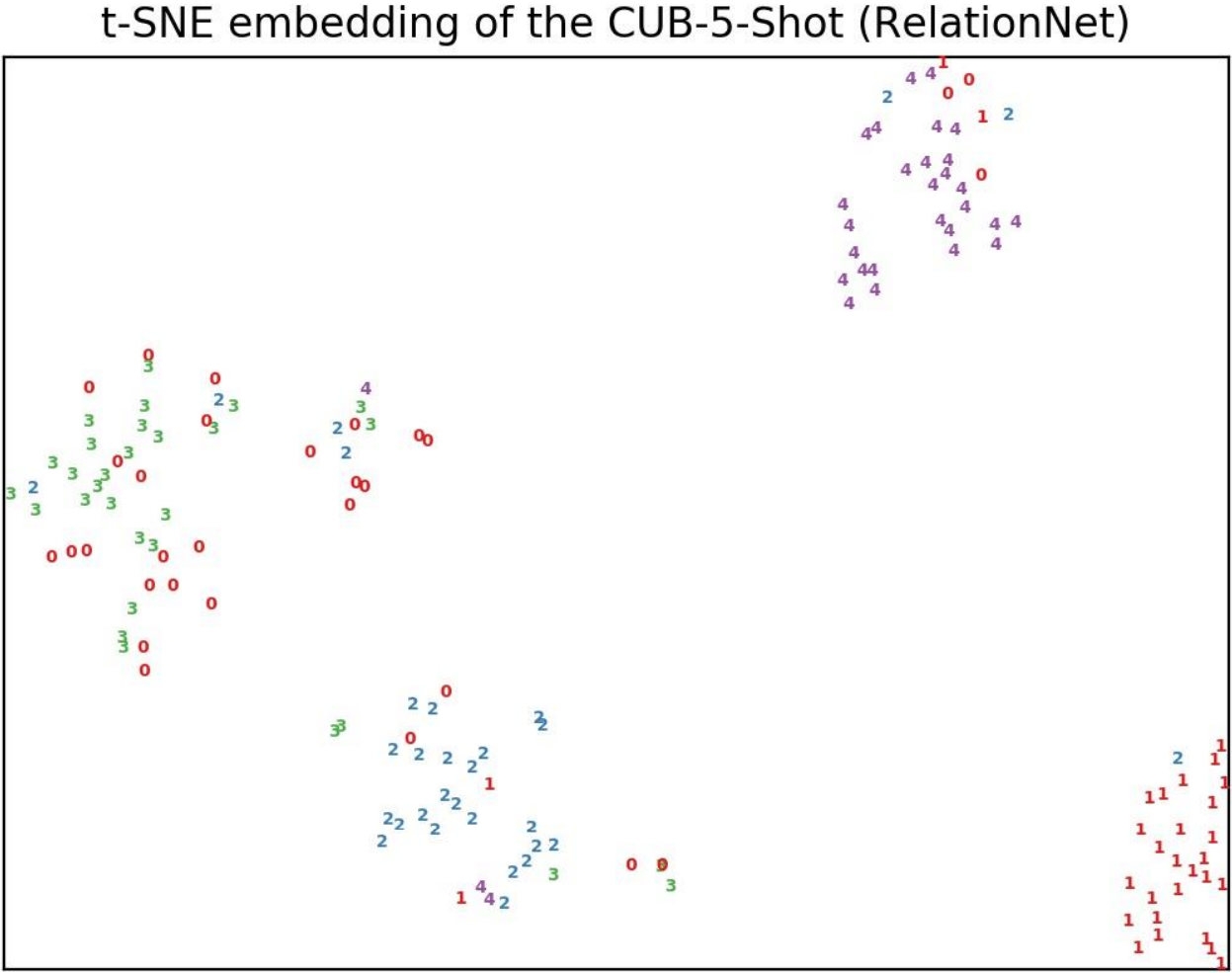} \label{sub01}}
	\subfloat[By LRPABN-Dim-128] {\includegraphics[width=0.245\textwidth]{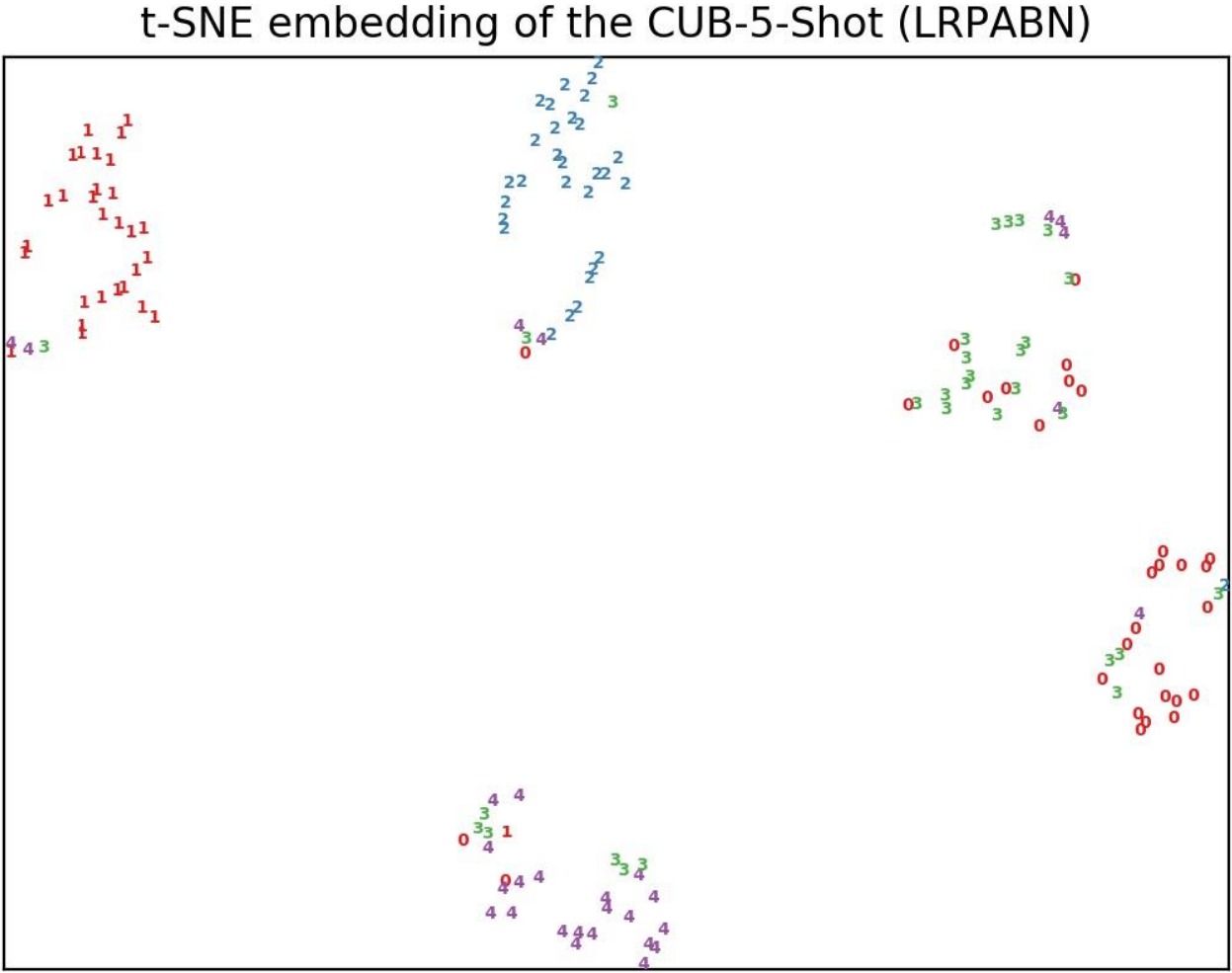} \label{sub02}}
	\subfloat[By PABN+] {\includegraphics[width=0.245\textwidth]{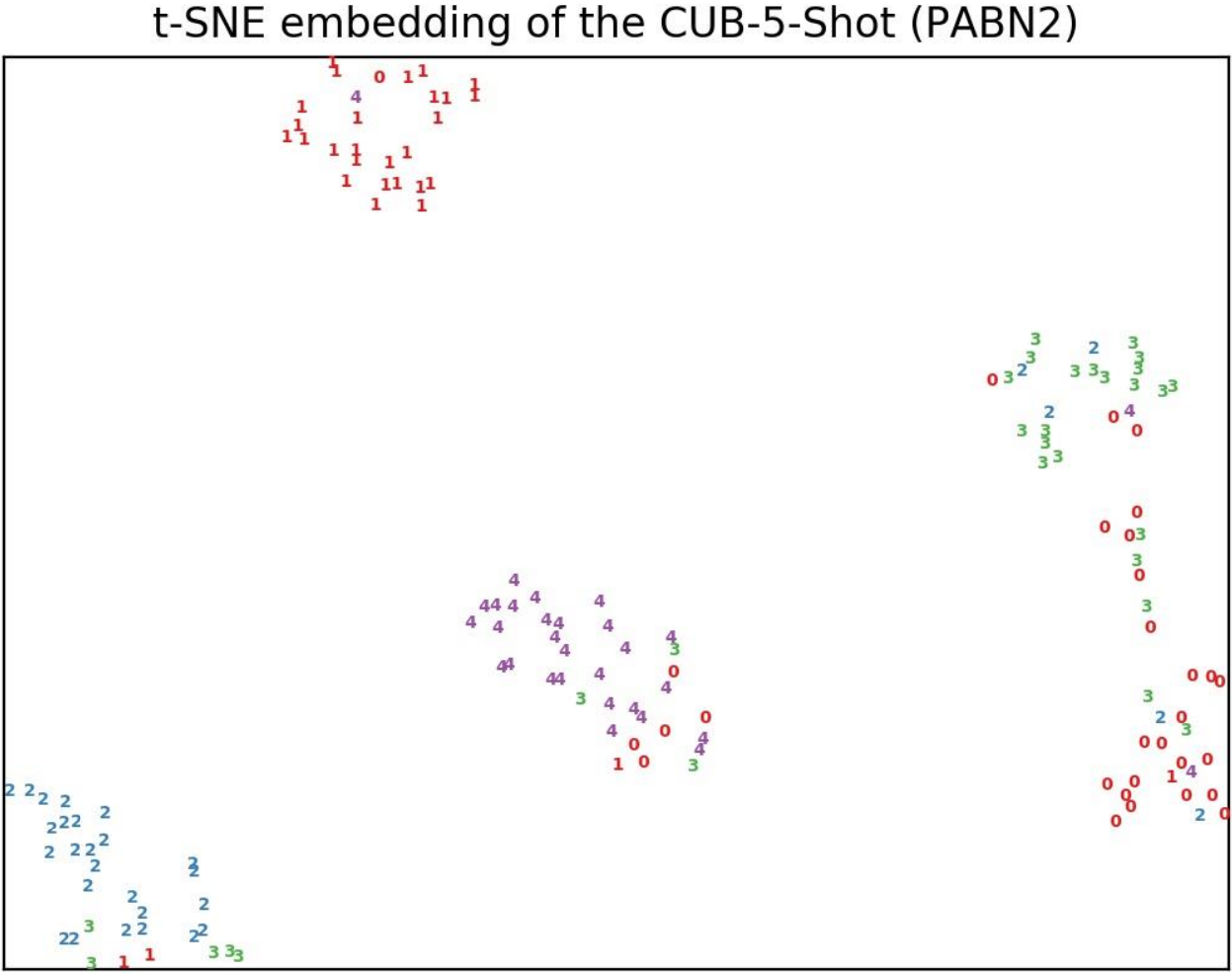} \label{sub03}}
	\subfloat[By LRPABN-Dim-512] {\includegraphics[width=0.245\textwidth]{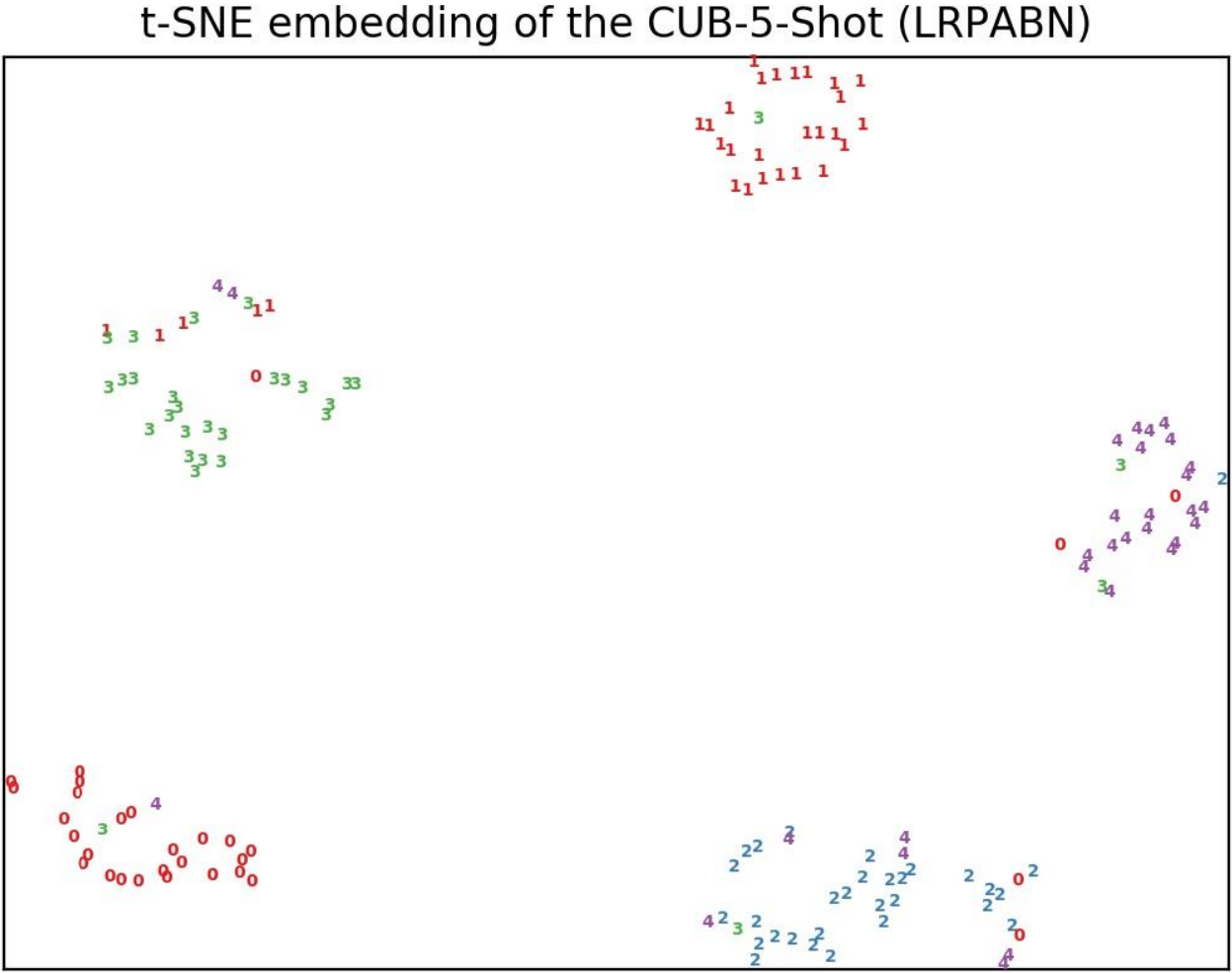} \label{sub04}}
	\caption{Visualization of the comparative feature generated by different fusion mechanism in 2D space using t-SNE~\cite{maaten2008visualizing}. Each dot represents a query image that is numeric and marked with different colors according to the real labels. For each class, we randomly select thirty query images to conduct this experiment. The visualization is based on the CUB data set under the 5-way-5-shot setting. (a) shows the result conducted by RelationNet, (b) shows the result conducted by LRPABN$_{cpt}$, and the dimension of the comparative bilinear feature is 128, denoted as LRPABN-Dim-128, (c) shows the result conducted by PABN+$_{cpt}$ model and (d) shows the result conducted by LRPABN$_{cpt}$, and the dimension of the comparative bilinear feature is 512, denoted as LRPABN-Dim-512.}
	\label{figTSNE}
\end{figure*}

Table~\ref{tab3} presents the average accuracies of different models on the novel classes of the fine-grained data sets. Both the one-shot and five-shot classification results are reported. As the table shows, the proposed LRPABN models get steadily and notably improvements on almost all fine-grained data sets for different experimental settings. More detailed, compared with CovaMNet, our proposed models achieve plainly growth performances on CUB Birds, CARS, and NABirds data sets on both one-shot and five-shot setting. Especially for NABirds data, the LRPABN$_{cpt}$ obtains $7.70\%$ and $5.99\%$ gain over CovaMNet for one-shot and five-shot setting, respectively. 
These results again firmly prove that the proposed pairwise bilinear pooling is superior compared to the self-bilinear pooling operation. Meanwhile, the feature alignment layer further boosts the final performance.  

For the comparisons against the DN4 method, from Table~\ref{tab3}, LRPABN models obtain the highest accuracy on one-shot setting on CUB Birds, CARS, NABirds data sets, and get second best results on DOGS data, where DN4 performs poorly in one-shot tasks on almost all data sets. 
For the five-shot setting, DN4 achieves the highest classification accuracy on all four data sets, while LRPABN$_{cpt}$ achieves the second-highest performance on CUB Birds, CARS, and NABirds. We are surprised to observe that in the one-shot-five-way task on the NABirds data, LRPABN$_{cpt}$ gains $15.92\%$ over DN4. Nevertheless, DN4 gets $15.36\%$ boosts over LRPABN$_{cpt}$ under the five-shot-five-way setting on the CARS data set.
That is, the proposed LRPABN method holds a tremendous advantage over DN4 for one-shot classification tasks but slightly inferior to DN4 for five-shot classification. 

The reason for this is that DN4 uses a deep nearest neighbor neural network to search the optimal local features in the support set as the support classes' feature for a given query image. For the target query features (\textit{e.g.,} a set of local features), the algorithm selects the top k nearest local features in the whole support data set according to the cosine similarity between query local features and support local features. That is, the more image in the support classes, the better the class feature will be generated. Thus, for five-shot classification, the DN4 outperforms LRPABN, where under the one-shot setting, DN4 has smaller support features to extract a good representation of the class feature. 
\textcolor{black}{More importantly, our model is more efficient than DN4. Specifically, under the $C$-way-$K$-shot setting, in the inference stage, for each query image, DN4 has $ h^2 \times w^2 \times K \times C \times O_{cos}$ computations to predict its label, while LRPABN only needs $ h \times w \times C \times O_{comp}$ computations. $h$ and $w$ denote the height and width of the feature map, $O_{cos}$ means the cosine similarity computation used in DN4, and $O_{comp}$ represents the comparator computation in LRPABN. Since $h \times w \times K \times O_{cos} \gg O_{comp}$, DN4 is much slower than LRBPAN during both training and testing, as seen from Table~\ref{ABS}, DN4 costs 15.20 $\times 10^{-3}$ s for each query, while LRPABN only needs 2.23 $\times 10^{-3}$ s, which is approximately seven times faster. Moreover, without considering the computation load, our initial low-rank pairwise bilinear model PABN$_{new}$ (Equation (\ref{eq41})) can also achieve the comparable performance against DN4 under both one-shot and five-shot setting, \textit{i.e.,} 78.87\% for PABN$_{new}$ compared to 79.64\% for DN4 under the five-shot settings.} 
\textcolor{black}{On the other hand, in many practical scenarios, such as endangered species protection, we may only get a one-labeled sample. With higher accuracy under the one-shot setting, our method can achieve more reliable performances compared to DN4 under such circumstances. It indicates the practical value of our models.}
Considering the proposed LRPABN summing the image features in each category as the class feature, how to generate a good representation of category would further improve the classification performance of our methods.

The classification examples of LRPABN, PABN+, and RelationNet models are shown in Figure \ref{Imshow}. We select LRPABN$_{cpt}$ and PABN+$_{cpt}$ as the representative of LRPABN and PABN+ approaches. To investigate the low-rank approximation, we set low-rank comparative feature dimensions as 512 and 128 for LRPABN-512 and LRPABN-128 models separately. By sending a fixed testing batch through the model, which consists of one support sample and five query samples for each of five classes, the prediction of LRPABN-512 only contains six mislabels in the entire 25 queries, while the prediction of LRPABN-128, PABN+ and RelationNet have 7, 8 and 10 wrong labels separately. That validates the effectiveness of the LRPABN models.
We also find that in some classes like Nighthawk and Harris Sparrow, the high intra-variance and low inter-variance confuse all the models.

\subsection{Ablation studies}
\textcolor{black}{Following the data split used in \cite{wei2018piecewise,huang2019compare}, we conduct several experiments to investigate the different components of the proposed model, the experimental results are shown in Table~\ref{ABS}. We analysis our methods from various aspects:}

\textcolor{black}{\textbf{Low-Rank Pairwise Bilinear Pooling:} First, we replace previous pairwise bilinear pooling (Equation~(\ref{eq4})) with Equation~(\ref{eq41}) as PABN$_{new}$. As seen in Table~\ref{ABS}, PABN$_{new}$ outperforms PABN$_{cpt}$ on both 1-shot and 5-shot tasks with a lower dimension, which indicates the effectiveness of our proposed initial Low-Rank pairwise pooling (Equation~(\ref{eq41})). However, using Equation~(\ref{eq41}), the model needs to learn a $n \times c \times c$ transformation tensor $\mathcal{W}$ (discussed in Section~\ref{LRPABN-1}), which significantly increases the model size and inference time. Thus, we employ Equation~(\ref{eq42}) to approximate the transformation tensor as LRPABN. We observe that this approximation achieves superior performance against our previous PABN$_{cpt}$ with a reduced model size as well as a shorter bilinear feature dimension. Specifically, as observed in Table~\ref{ABS}, the proposed LRPABN costs 2.23 $\times 10^{-3}$ s to identify a query image with a 213K model size, while the previous ICME model PABN$_{cpt}$ requires 8.65 $\times 10^{-3}$ s and 375K parameters. 
Moreover, the inference time of LRPABN is 2.23$\times 10^{-3}$ s, while PABN$_{new}$ costs 78.40 $\times 10^{-3}$ s for each query image. That is, our final low-rank pairwise pooling model LRPABN is more advanced than previous PABN models and much  more efficient than PABN$_{new}$ model.}

\textcolor{black}{\textbf{Alignment Mechanism:} To investigate the effectiveness of the proposed alignment mechanism. We compare PABN$_{cpt}$ and PABN+$_{cpt}$. Besides, 
we adopt the proposed alignment loss $Align_{loss_{2}}$ in Equation~(\ref{eq6}) into LRPABN as LRPABN$_{only\_cpt}$. As seen from Table~\ref{ABS}, cooperating with the position transform function \textbf{T}, PABN+$_{cpt}$ and LRPABN$_{cpt}$ outperform PABN$_{cpt}$ and LRPABN$_{only\_cpt}$, respectively. For instance, under the 5-shot setting, the classification accuracy of PABN+$_{cpt}$ is 77.19\% compared to 76.81\% of PABN$_{cpt}$. }

\textcolor{black}{\textbf{Input Image Size:} It is reported that a higher resolution of the input image can capture a more discriminative feature for Fine-grained classification~\cite{Lin_2015_ICCV,Cui_2017_CVPR,Li_2018_CVPR}. However, few-shot learning models~\cite{vinyals2016matching,snell2017prototypical,Sung_2018_CVPR} usually adopt a low input resolution, e.g., 84 $\times$ 84. For a fairness comparison with generic few-shot learning approaches, in Section~\ref{setup}, we set the input image size to 84 $\times$ 84. To further investigate the affects of input size, we follow \cite{wei2018piecewise} to replace the shallow embedding network Conv4~\cite{vinyals2016matching,snell2017prototypical,Sung_2018_CVPR} with AlexNet~\cite{krizhevsky2012imagenet} as LRPABN$_{cpt}$:AlexNet. Moreover, we choose two resolutions for the input images, which are widely used in Fine-grained classification. As Table~\ref{size} shows, with AlexNet, a higher resolution 448 $\times$ 448 brings a significant performance boost compared to lower input size 224 $\times$ 224, which validates that a higher input resolution can generate a more subtle comparative feature for FSFG.
We also observe that the accuracy of the AlexNet-based methods performs worse than Conv4-based methods. A high input resolution always accompanied by a deep embedding network like AlexNet to extract the informative feature. However, training a deeper embedding network with limited labeled samples is easier to lead the over-fitting problem. 
}


\textbf{Bilinear Feature Dim:}
For the feature dimension selection, we change the number of dimensions as 16, 32, 64, 128, 256, 512, 1024, and 2048 for both 1-shot and 5-shot classification tasks on CUB Birds data. The model we used for this experiment is LRPABN$_{cpt}$. The results are shown in Figure~\ref{figdim}, it can be observed that as the feature dimension gets large, the test accuracy gradually improves to a peak first, then it goes through a drastic drop. For the 1-shot setting, the performance changes smoothly when the dimension is below 1024. For the 5-shot task, the variation of performance is relatively oscillatory, yet it can grow fast and steadily, with the dimension increasing. Moreover, we find that even with a very compact low-rank approximation (\textit{i.e.,} the dimension is 16), the model can still achieve a decent classification performance, which fatherly verifies the stability of the proposed method. When the dimension goes too large, the model performs poorly, and this may be caused by the increased complexity of the framework can not model the data distribution well with few training samples. \textcolor{black}{As \cite{Gao_2016_CVPR} discussed, for self-bilinear features, less than 5\% of dimensions are informative. For FSFG, the best feature dimensions for LRPABN are 256 and 512 in the experiments, which are around 5\% to 10\% of the entire self-bilinear feature dimension.}

\textbf{t-SNE visualization:}
The t-SNE~\cite{maaten2008visualizing} visualization for different comparative features is presented in Figure~\ref{figTSNE}. We randomly select five support images and thirty query images per category from CUB Birds data to conduct the five-way-five-shot tasks. The original comparative feature dimension of RelationNet is $128\times3\times3$. We use the convolved feature before the first fully-connected layer in classifier as the final comparative feature with dimension size 576. The comparative feature of PABN+ is $64\times64=4096$, and we choose LRPABN$_{cpt}$ with comparative dimension 128 and 512 separately (denoted as LRPABN-Dim-128 and LRPABN-Dim-512) for comparison. As the figure shows, the learned LRPABN-Dim-512 feature, which can be grouped into five classes correctly, outperforms others, the discriminative performance of LRPABN-Dim-128 and PABN+ are similar, which outperform RelationNet' feature. The intuitive visualization results among the above methods again validate the superior capacity of the proposed low-rank pairwise bilinear features for FSFG tasks.

\section{Conclusion}\label{conc}
In this paper, we propose a novel few-shot fine-grained image classification method, which is inspired by the advanced information processing ability of human beings. The main contribution is the low-rank pairwise bilinear pooling operation, which extracts the second-order comparative features for the pair of support images and query images. Moreover, to get a more precise comparative feature, we propose an effective feature alignment mechanism to match the embedded support image features with query ones. Through comprehensive experiments on four fine-grained datasets, we verify the effectiveness of the proposed method. \textcolor{black}{
As the future work, we will investigate more sophisticated alignment mechanisms that applies the feature transformation to support and query images jointly.}


%





\ifCLASSOPTIONcaptionsoff
  \newpage
\fi

\bibliographystyle{IEEEtran}
\bibliography{ieeeTrans}






\end{document}